\documentclass[10pt,twocolumn,letterpaper]{article}

\usepackage{cvpr}              % To produce the CAMERA-READY version
\usepackage[dvipsnames]{xcolor}

\definecolor{cvprblue}{rgb}{0.21,0.49,0.74}
\usepackage[pagebackref,breaklinks,colorlinks,citecolor=cvprblue]{hyperref}

\usepackage{graphicx}
\usepackage{fdsymbol}
\usepackage{lipsum}
\usepackage{amsmath}
\usepackage{amssymb}
\usepackage{booktabs}
\usepackage{listings}
\usepackage[export]{adjustbox}
\usepackage{epsfig}
\usepackage{algorithm}
\usepackage{algpseudocode}
\usepackage{enumitem}
\usepackage{colortbl}
\usepackage{url}

\usepackage{soul}
\usepackage{multirow}
\usepackage{capt-of}
\usepackage{arydshln}
\usepackage{blkarray, bigstrut} %
\usepackage{mathtools}

\newcommand*{\img}[1]{%
    \raisebox{-.15\baselineskip}{%
        \includegraphics[
        height=1\baselineskip,
        width=1\baselineskip,
        keepaspectratio,
        ]{#1}%
    }%
}
\definecolor{codegreen}{rgb}{0,0.6,0}
\definecolor{codegray}{rgb}{0.5,0.5,0.5}
\definecolor{codepurple}{rgb}{0.58,0,0.82}
\definecolor{backcolour}{rgb}{0.95,0.95,0.92}
\definecolor{mygray}{RGB}{218,218,218}
\definecolor{ftcolor}{RGB}{255,204,246}
\definecolor{zscolor}{RGB}{191,248,255}
\definecolor{c_green}{RGB}{180,255,184}
\definecolor{h_orange}{RGB}{255,224,208}
\definecolor{qa_blue}{RGB}{183,247,255}
\definecolor{qa_pink}{RGB}{255,193,246}
\definecolor{blue_ref}{RGB}{54,125,189}

\lstdefinestyle{mystyle}{
  backgroundcolor=\color{backcolour}, commentstyle=\color{codegreen},
  keywordstyle=\color{magenta},
  numberstyle=\tiny\color{codegray},
  stringstyle=\color{codepurple},
  basicstyle=\ttfamily\footnotesize,
  breakatwhitespace=false,         
  breaklines=true,                 
  captionpos=b,                    
  keepspaces=true,                 
  numbers=left,                    
  numbersep=5pt,                  
  showspaces=false,                
  showstringspaces=false,
  showtabs=false,                  
  tabsize=2
}

\newcommand{\modelname}{\textbf{V$^2$Dial\,}}
\definecolor{purple}{RGB}{255,175,251}
\definecolor{myblue}{RGB}{157,242,255}
\definecolor{myorange}{RGB}{255,227,200}
\definecolor{violet}{RGB}{101,0,174}
\definecolor{pink}{RGB}{255,121,248}
\definecolor{skyblue}{RGB}{53,172,255}
\definecolor{green}{RGB}{0,174,10}
\definecolor{orange}{RGB}{255,138,82}
\definecolor{blue_ref}{RGB}{54,125,189}

\DeclareMathOperator*{\argmax}{arg\,max}

\newcommand{\rom}[1]{\uppercase\expandafter{\romannumeral #1\relax}}
\usepackage{pifont}% http://ctan.org/pkg/pifont
\newcommand{\cmark}{\ding{51}}%
\newcommand{\xmark}{\ding{55}}%

\title{\modelname \img{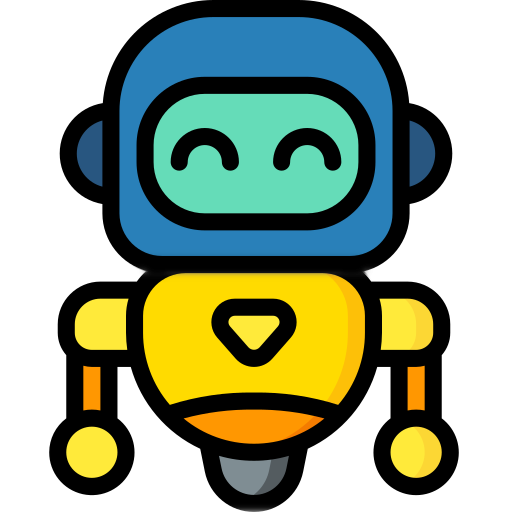}: Unification of \underline{V}ideo and \underline{V}isual \underline{Dial}og via Multimodal Experts}

\def\paperID{16269} % *** Enter the Paper ID here
\def\confName{CVPR}
\def\confYear{2025}

\author{Adnen Abdessaied\textsuperscript{ \img{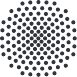}} \quad
Anna Rohrbach\textsuperscript{\img{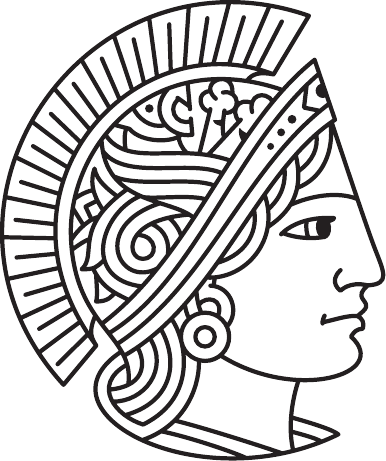} \img{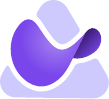}} \quad
Marcus Rohrbach\textsuperscript{\img{figures/logo_tudarmstadt.pdf} \img{figures/logo_hessianai.pdf}} \quad 
Andreas Bulling\textsuperscript{\img{figures/logo_stuttgart.pdf}} \vspace{0.3em} \\
{\normalsize \textsuperscript{\img{figures/logo_stuttgart.pdf}}University of Stuttgart, Germany} \quad
{\normalsize \textsuperscript{\img{figures/logo_tudarmstadt.pdf}}TU Darmstadt, Germany} \quad
{\normalsize \textsuperscript{\img{figures/logo_hessianai.pdf}}hessian.AI, Germany} \\
{\normalsize \url{https://www.collaborative-ai.org/publications/abdessaied25_cvpr/}}
}

\begin{document}
\maketitle

\begin{abstract}
We present \modelname -- a novel expert-based model specifically geared towards simultaneously handling image and video input data for multimodal conversational tasks.
Current multimodal models primarily focus on simpler tasks (e.g., VQA, VideoQA, video-text retrieval) and often neglect the more challenging conversational counterparts, such as video and visual/image dialog.
Moreover, works on both conversational tasks evolved separately from each other despite their apparent similarities, limiting their applicability potential.
To this end, we propose to unify both tasks using a single model that for the first time jointly learns the spatial and temporal features of images and videos by routing them through dedicated experts and aligns them using matching and contrastive learning techniques.
Furthermore, we systemically study the domain shift between the two tasks by investigating whether and to what extent these seemingly related tasks can mutually benefit from their respective training data.
Extensive evaluations on the widely used video and visual dialog datasets of AVSD and VisDial show that our model achieves new state-of-the-art results across four benchmarks both in zero-shot and fine-tuning settings.
\end{abstract}
    
\section{Introduction}
\begin{figure}[!t]
    \begin{minipage}{1\linewidth}
        \centering
        \scalebox{0.9}[0.9]{
            \includegraphics[width=\textwidth]{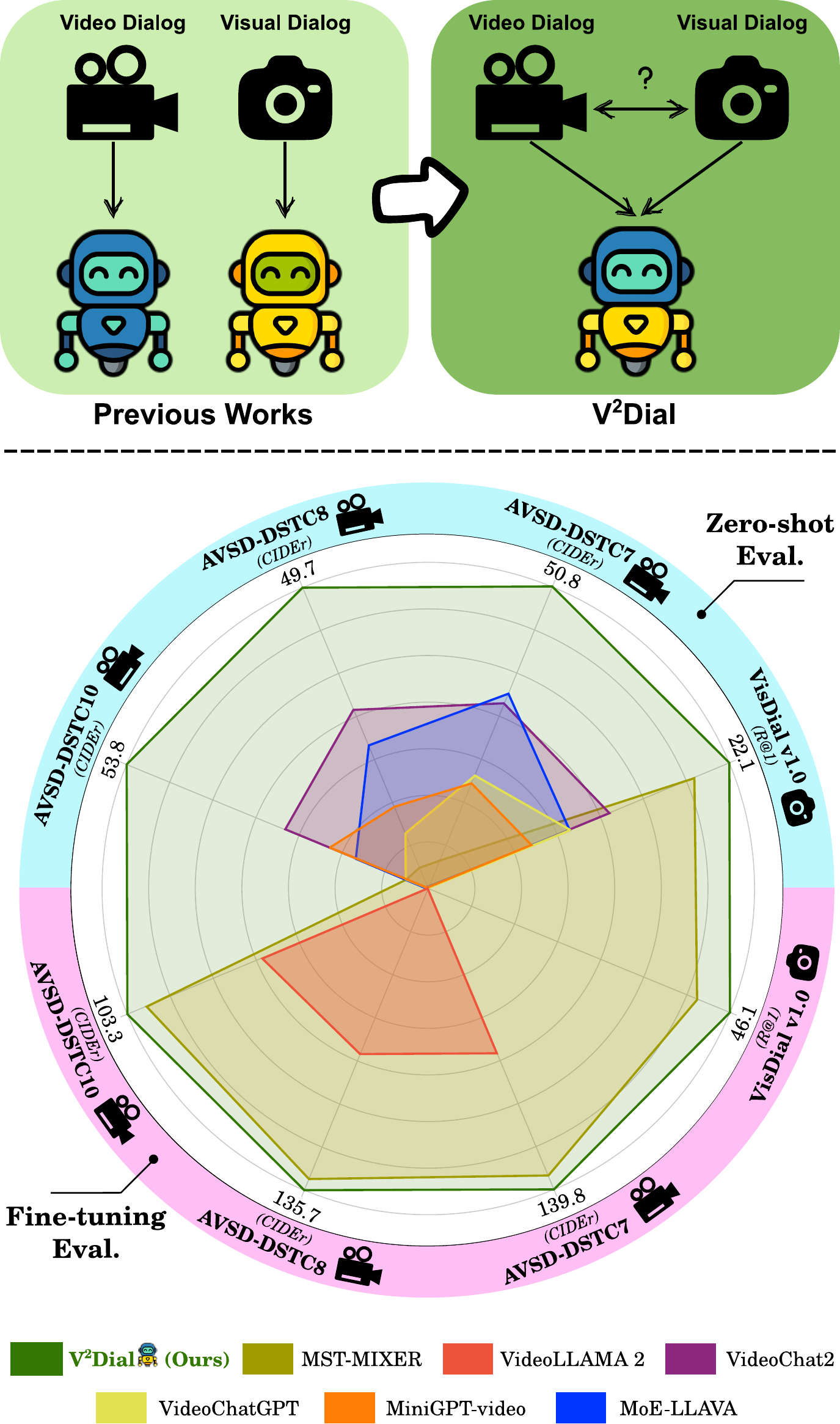}
        }
        \caption{\modelname \img{figures/robot_mixed.png} uses multimodal experts and outperforms state-of-the-art methods on both video and visual dialog in zero-shot and fine-tuning evaluation settings.  
        }
        \label{fig:teaser}
    \end{minipage}
\end{figure}
Enabled by the availability of large-scale training data \cite{cc12m,howto100m,webvid} and advances in model design \cite{transformer,vit,st,bao2022vlmo,llava}, the field of vision-and-language learning saw unprecedented success in recent years.
However, current multimodal foundational models \cite{llava,flamingo,flava,foreign_language,blip2} still mainly focus on single-round tasks (e.g., VQA \cite{vqa}, VideoQA \cite{videoqa}, video-text and text-video retrieval \cite{msrvtt}).
In contrast, the significantly more challenging conversational tasks, such as visual \cite{visdial,abdessaied2022neuro} and video dialog \cite{avsd}, received considerably less attention.
Furthermore, methods for these different tasks have advanced independently of each other despite the apparent structural similarities between them.
They both operate on a visual input (i.e. an image or video), a short visual description (caption), and a dialog history composed of previous question-answer pairs.
On the one hand, visual dialog models \cite{vdgr,Nguyen2020,Chen2022,GOG,wang2022unified} have been primarily trained to rank a list of candidate answers using a Next Sentence Prediction (NSP) head similar to BERT \cite{bert} and negative sampling.
Thus, they are benchmarked using retrieval metrics such as recall (R@k) and normalized discounted cumulative gain (NDCG).
In contrast, video dialog models \cite{mst_mixer,abdessaied_olvit,le2021learning,pham2022video,mrlv,tham,dmcf,video_llama2} are trained to auto-regressively predict the next answer token using teacher forcing \cite{williams1989learning} and are evaluated using language generation metrics.

In this work, we mitigate the shortcomings of current dialog systems by proposing \modelname -- a novel multimodal expert-based model capable of unifying video and visual dialog tasks without any architectural changes.
Specifically, we train dedicated multimodal expert layers that separately process the features of each input modality and learn how to align them using matching and contrastive learning techniques.  
A key novelty of our approach is that we use dedicated experts to jointly learn the spatial and temporal features of images and videos by routing them through the appropriate experts.
Then, we couple these layers with a pre-trained LLM to align their hidden states.
Thanks to its modularity, our model can efficiently tackle image and video input data simultaneously and seamlessly train on both data types.
In summary, our contributions are three-fold:
\begin{itemize}[wide,parsep=0.1em,labelindent=0.0cm,leftmargin=0.2cm]
    \item We propose \modelname -- a multimodal expert-based model that unifies visual and video dialog by simultaneously learning from image and video data.
    As a core novelty, it employs two experts to separately learn the spatial and temporal features of images and videos.
    \modelname outperforms state-of-the-art models in both zero-shot and fine-tuning settings (see \autoref{fig:teaser}).
    \item We are the first to systematically quantify the effect of domain shift between video and visual dialog tasks based on evaluations on the two widely used datasets of AVSD \cite{avsd} and VisDial \cite{visdial}.
    To this end, we propose an alternative ranking scheme that allows computing the VisDial retrieval metric for fully generative models and enables a fair comparison with previous works. 
    \item We are the first to evaluate AVSD in a zero-shot setting, which provides a more solid generalization evaluation of video dialog models compared to the fine-tuned setting.
    For this, we establish the first benchmark comparison of recent state-of-the-art multimodal models.
\end{itemize}

\section{Related Work}
\textbf{Visual \& Video Dialog.}
Modeled after human-human communication, visual and video dialog involve reasoning about a visual scene in the form of a video or an image through multiple question-answering rounds in natural language.
In comparison to their single-round counterparts, VQA \cite{vqa} and VideoQA \cite{videoqa}, dialog models need to additionally reason about the previous dialog history together with the visual grounding and the current question to be able to answer it efficiently.
The best performing visual dialog models \cite{vdgr, wang2022unified, murahari2020large, vd_pcr} leverage pre-trained VLMs and are trained using an NSP head, negative sampling, and binary classification loss.
At test time, for each question, the candidate answers are ranked based on their respective NSP scores to compute the retrieval metrics.
Although some work \cite{vdbert, Chen2022, GOG} claim to train generative visual dialog models, they do so by providing a generative mask where each token can only attend to its left tokens. However, they are trained using the NSP head like the discriminative models.
However, this training approach is limiting and suboptimal for a unifying model.
Thus, we advocate for a fully generative training paradigm and adapt the ranking scheme of VisDial answers to cater to modern generative models.
\begin{figure*}[!t]
    \begin{minipage}{1\linewidth}
        \centering
        \scalebox{0.98}[0.98]{
            \includegraphics[width=\textwidth]{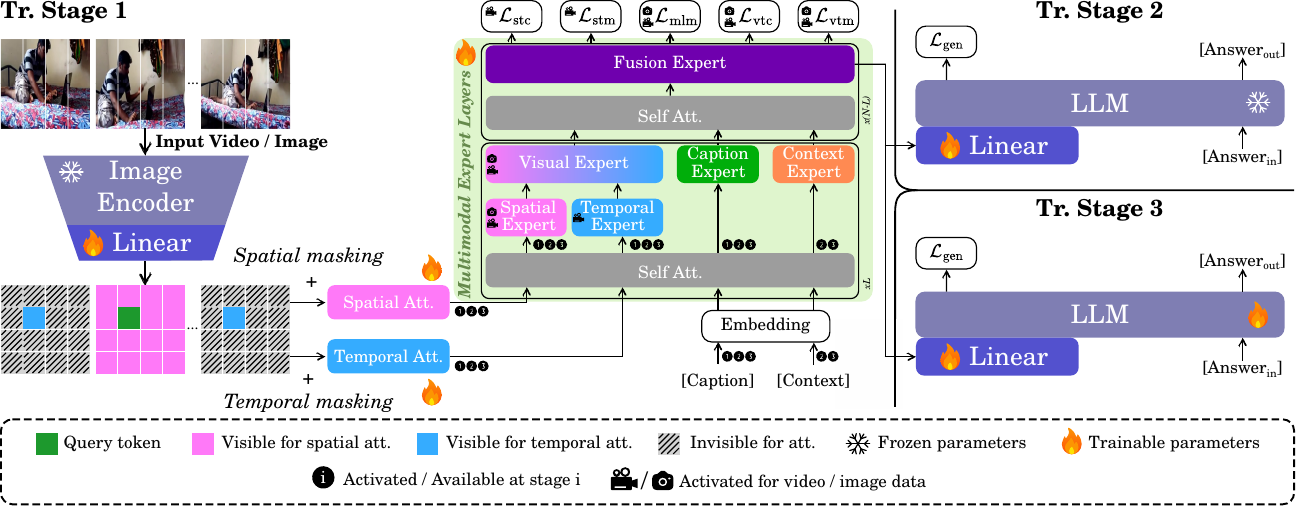}
        }
        \caption{\textbf{Architectural overview of} \modelname \img{figures/robot_mixed.png}. We adopt a training strategy composed of three stages. \textit{First}, we only train the multimodal expert layers using spatial-temporal and video/image text 
        matching losses ($\mathcal{L}_\textrm{stm}, \mathcal{L}_\textrm{vtm}$), spatial-temporal and video/image contrastive learning losses ($\mathcal{L}_\textrm{stc}, \mathcal{L}_\textrm{vtc}$), and masked language modeling loss ($\mathcal{L}_\textrm{mlm}$).
        \textit{Second}, we couple the expert layers with a frozen pre-trained LLM end-to-end, using a generative loss $\mathcal{L}_\textrm{gen}$ to align their hidden representations.
        \textit{Finally}, we additionally fine-tune the LLM weights on the downstream benchmarks.
        Each expert is a feed-forward network (FFN) composed of two fully connected layers.
        }
        \label{fig:main_method}
    \end{minipage}
\end{figure*}

In contrast, works on video dialog follow a purely-generative training paradigm and achieved great success building on top of powerful pre-trained LLMs \cite{bart,gpt}.
For example, \cite{le-hoi-2020-video, 9376902} fine-tuned a LLM on AVSD and obtained performance boosts.
More recent works \cite{le2021learning, mst_mixer} combined LLMs with GNNs and pushed the state-of-the-art results even further.
Others \cite{tham} introduced a regularization loss to mitigate hallucination.
Although video dialog emerged as a natural extension to visual dialog with apparent data structure similarities, research on both tasks evolved separately.
To this end, we propose a unifying model that can simultaneously learn both tasks without any architectural modifications and \textit{for the first time}; systemically study the effect of domain shift between both tasks using the AVSD and VisDial v1.0 datasets.\\

\noindent\textbf{Multimodal Expert-based Training.}
Enhancing models with expert-based training has shown promising potential in boosting performance while maintaining computational efficiency \cite{glam,llama_moe,zhong2024lory}.
Some works \cite{bao2022vlmo, foreign_language} explored using single modality specific experts within a multimodal transformer architecture. Specifically, they used \textit{one} vision and \textit{one} language specific FFN after a shared multi-head self-attention block. 
Other works \cite{moe_llava,NEURIPS2022_3e67e84a} explored using multiple sparse modality-agnostic experts and trained them using soft-routers.
Our work is positioned at the middle ground of the previously mentioned research directions: We propose to use multiple hard-routed experts per modality to be able to capture more fine-grained features compared to a single expert or multiple modality agnostic experts.
Specifically, to the best of our knowledge, \modelname is the \textit{first} model that learns disentangled spatial and temporal features using two dedicated experts that jointly learn from image and video data.
In addition, we propose to deploy two separate language experts (for caption and context) in order to tackle the unique challenges of multimodal conversational tasks. 
\section{\modelname}
\subsection{Joint Problem Formulation}
We use a fully generative formulation to unify both video and visual dialog tasks.
Specifically, given visual input  \texttt{V} (video/image), a corresponding visual description (caption \texttt{C}), a dialog history \texttt{H$_r$} = \{(\texttt{Q$_1$}, \texttt{A$_1$}), ..., (\texttt{Q$_{r-1}$}, \texttt{A$_{r-1}$})\} comprised of the previous question-answer pairs 
\{(\texttt{Q$_i$},\texttt{A$_i$})\}$_{i=1}^{r-1}$ and the current question \texttt{Q}$_{r}$, a model is trained to auto-regressively predict a free-form answer \texttt{A}$_r$ at round $r$. 
Specifically, each answer token \texttt{a$_r^i$} satisfies 
\begin{equation}
    \texttt{a}_r^i = \displaystyle \argmax_{\texttt{a} \in \mathcal{V}} \left[ \mathbf{p}\left(\texttt{a} | \texttt{V}, \texttt{C}, \texttt{H}_\texttt{r} \texttt{Q}_\texttt{r}, \texttt{A}_\texttt{r}^{<\texttt{i}} \right)\right],%      
\end{equation}
where $\texttt{A}_{r}^{<i}$  denotes the previously predicted answer tokens and  $\mathcal{V}$ the vocabulary.
In the rest, we use \textit{``context''} to refer to the concatenation of the history $\texttt{H}_\texttt{r}$ and the question $\texttt{Q}_\texttt{r}$.

\subsection{Architecture}
\paragraph{Overview.}
As can be seen from \autoref{fig:main_method}, our model takes an image/video $\texttt{V} \in \mathbb{R}^{F\times 3 \times H \times W}$ as input, where $F$ is the number of frames and is set to \textit{one} for images, and $(H, W)$ is the re-sized resolution.
Then it processes every frame using a pre-trained EVA-CLIP \cite{eva_clip} Image Encoder and concatenates every four spatially adjacent visual patches into a single one.
Then, a linear layer maps each visual token into a lower dimensional vector $\mathbf{v}$ of dimension $D$ to obtain
\begin{equation}
    \mathbf{V} = \begin{bmatrix}
        \mathbf{v}_1^1  & \mathbf{v}_1^2  & \cdots & \mathbf{v}_1^F  \\
        \vdots & \vdots & \ddots & \vdots \\
        \mathbf{v}_P^1  & \mathbf{v}_P^2  & \cdots & \mathbf{v}_P^F  \\
    \end{bmatrix}  \in \mathbb{R}^{F\times P \times D},
\end{equation}
where $P = \frac{1}{4}\frac{H \times W}{14^2}$ and $D$ denote the visual input length and the joint hidden dimension, respectively.
Thereafter, in stark contrast to previous works \cite{vindlu,space_time_att} that performed spatial and temporal attention in series, our model \textit{separately} performs these operations using the masks $\mathbf{M}^{\textrm{spa}}$ and $\mathbf{M}^{\textrm{tmp}}$ as shown in \autoref{fig:main_method} on the visual features $V$ to obtain
\begin{align}
    &\mathbf{V}^{\textrm{spa}} = \mathrm{SA}(\mathbf{V}, \mathbf{M}^{\textrm{spa}}) \in \mathbb{R}^{(FP)\times D}\\
    &\mathbf{V}^{\textrm{tmp}} = \mathrm{SA}(\mathbf{V}, \mathbf{M}^{\textrm{tmp}})\in \mathbb{R}^{(FP)\times D}\\
    &\mathbf{M}^{\textrm{spa}}_{m,n}(\mathbf{v}_i^j) = \delta_{nj},\,\, \mathbf{M}^{\textrm{tmp}}_{m,n}(\mathbf{v}_i^j) = \delta_{mi}
\end{align}
where $\mathrm{SA}$ and $\delta$ denote self-attention and Kronecker delta.

Subsequently, the textual input in the form of a caption and a context is processed by an embedding layer to obtain $ \mathbf{T}^{\textrm{cap/ctx}} \in \mathbb{R}^{N_{\textrm{cap/ctx}}\times D}$, where $N_{\textrm{cap}}$ and $N_{\textrm{ctx}}$ are the respective lengths of the caption and context.
These visual and textual features form the initial input to the multimodal expert layers which are pre-trained using a combination of matching, contrastive learning, and masked language modeling losses.
Finally, they are coupled with a pre-trained LLM and are fine-tuned end-to-end using a generative loss.

\begin{figure*}[!t]
    \begin{minipage}{1\linewidth}
        \centering
        \scalebox{0.94}[.94]{
            \includegraphics[width=\textwidth]{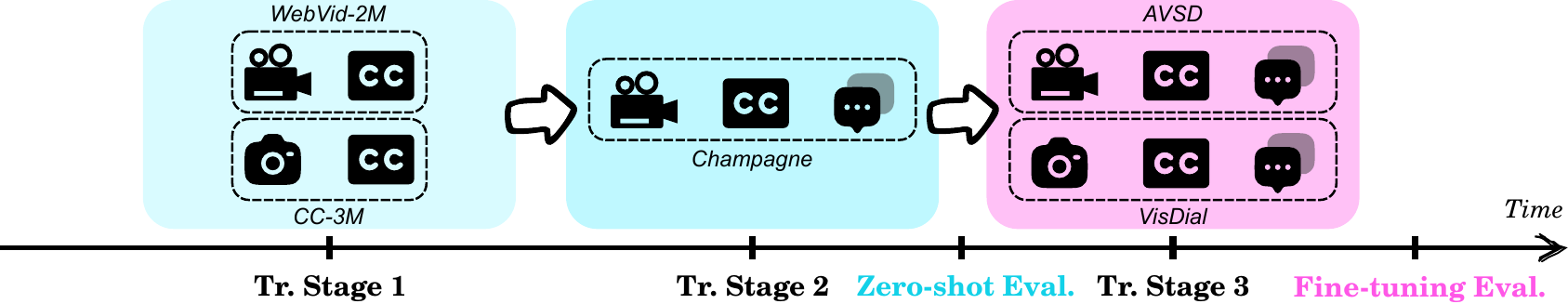}
        }
        \caption{\textbf{Overview of the training and evaluation pipeline of} \modelname \img{figures/robot_mixed.png}. 
        We show the different datasets used to train our model at each stage. Evaluations are conducted on the most popular video and visual dialog datasets of AVSD and VisDial, respectively. (\img{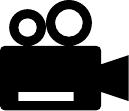} = video data, \img{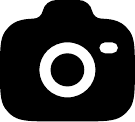} = image data, \img{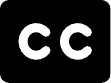} = closed / visual captioning data, \img{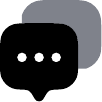} = dialog data).
        }
        \label{fig:pipeline}
    \end{minipage}
\end{figure*}

\paragraph{Multimodal Expert Layers.} These consist of $N$ layers of stacked multi-head self-attention with layer normalization (SA), and \textit{several} modality-specific and \textit{one} modality-agnostic feed-forward networks that we refer to as \textit{experts}.
As shown in \autoref{fig:main_method}, we propose to use a set of \textit{six} experts denoted as $\{\mathcal{E}_*\}$: \textit{three} of which are vision-specific and \textit{two} are language-specific and are activated in the first $L$ layers.
The \textit{remaining} expert $\mathcal{E}_\textrm{fus}$ is the fusion expert and is only activated in the last $(N-L)$ layers and operates on the concatenation of all available modalities (\autoref{eq:fusion_exp}).
To the best of our knowledge, we propose for the first time to learn the spatial and temporal features using dedicated experts (i.e., the spatial $\mathcal{E}_\textrm{spa}$ and temporal $\mathcal{E}_\textrm{tmp}$ experts, respectively) as shown in \autoref{eq:spa_tmp_exp}.
This allows our model to unify video and visual dialog by jointly learning from image and video data.
The visual expert $\mathcal{E}_{\textrm{vis}}$ operates on top of the concatenation of $\mathcal{E}_\textrm{spa}$ and $\mathcal{E}_\textrm{tmp}$ to learn a joint spatial-temporal video representation (\autoref{eq:vis_exp}).
Similarly, the textual experts $\mathcal{E}_\textrm{cap}$ and $\mathcal{E}_\textrm{ctx}$ operate on the caption and context embeddings $\mathbf{T}^\mathrm{cap}$ and $\mathbf{T}^\mathrm{ctx}$ (\autoref{eq:text_exp}).
As seen in \autoref{tab:input_to_expert_layers}, the availability of the multimodal features depends on the visual input type (i.e., videos vs images) and the training stage.
However, without the loss of generality, we can formulate the multimodal expert layers as follows:
\begin{small}
    \begin{align}
    \mathbf{X}_0 &= [\mathbf{V}^\textrm{spa},\mathbf{V}^\textrm{tmp}, \mathbf{T}^{\textrm{cap}}, \mathbf{T}^{\textrm{ctx}}],\\
    \mathbf{\tilde{X}}_{l} &=  [\mathbf{\tilde{V}}^\textrm{spa}_{l},
    \mathbf{\tilde{V}}^\textrm{tmp}_{l},
    \mathbf{\tilde{T}}^{\textrm{cap}}_{l},
    \mathbf{\tilde{T}}^{\textrm{ctx}}_{l}] \\
     &=\mathrm{SA}(\mathbf{X}_{l-1})+\mathbf{X}_{l-1}\\
    \mathbf{X}_{l} &= \begin{cases}
        [
        \mathbf{V}^\textrm{vis}_{l}\color{black},
        \color{green}\mathbf{T}^{\textrm{cap}}_{l}\color{black}, \color{orange}\mathbf{T}^{\textrm{ctx}}_{l} \color{black}] & if \,\, 1 \leq l \leq L
         \label{eq:fusion_exp} \\ 
        \color{violet}{\mathcal{E}_\textrm{fus}(\mathbf{\tilde{X}}_{l}) + \mathbf{\tilde{X}}_{l}} & if \,\, L < l \leq N
    \end{cases},\\
    \mathbf{{V}}^\textrm{vis}_{l} &=\mathcal{E}_\textrm{vis}(\tilde{\mathbf{V}}^\textrm{vis}_l) + \tilde{\mathbf{V}}^\textrm{vis}_l,\,\,
    \tilde{\mathbf{V}}^\textrm{vis}_l \coloneqq [\color{pink}{\mathbf{{V}}^\textrm{spa}_{l}}\color{black}, \color{skyblue}{\mathbf{{V}}^\textrm{tmp}_{l}}\color{black}],\label{eq:vis_exp}\\
    \color{pink}\mathbf{V}^\textrm{spa}_{l}&\color{pink}=\mathcal{E}_{\textrm{spa}}(\mathbf{\tilde{V}}^\textrm{spa}_{l})+ \tilde{\mathbf{V}}^\textrm{spa}_{l}\color{black},\,\, \color{skyblue}\mathbf{{V}}^\textrm{tmp}_{l}  \color{skyblue}{=\mathcal{E}_{\textrm{tmp}} (\mathbf{\tilde{V}}^\textrm{tmp}_{l})} + \mathbf{\tilde{V}}^\textrm{tmp}_{l}\color{black},\label{eq:spa_tmp_exp}\\
    \color{green}\mathbf{T}^{\textrm{cap}}_{l}&\color{green}= \mathcal{E}_{\textrm{cap}}(\mathbf{\tilde{T}}^{\textrm{cap}}_{l})+ \mathbf{\tilde{T}}^\textrm{cap}_{l}\color{black},\,\, 
    \color{orange}\mathbf{T}^{\textrm{ctx}}_{l} = \mathcal{E}_{\textrm{ctx}}(\mathbf{\tilde{T}}^{\textrm{ctx}}_{l})+ \mathbf{\tilde{T}}^\textrm{ctx}_{l}\color{black}.\label{eq:text_exp}
\end{align}
\end{small}
When dealing with images and non-dialog data, we drop $\mathbf{{V}}^\textrm{tmp}_{l}$ and $\mathbf{T}^{\textrm{cap}}_{l}$ from the previous equations and deactivate the respective expert.
\subsection{Training}
\subsubsection{Stage 1} In the first stage, we only pre-train the multimodal expert layers, the vision encoder linear layer, and the spatial-temporal attention modules.
Since we are the first to suggest learning the spatial and temporal features of videos and images using dedicated experts, we propose to train our model using spatial-temporal contrastive learning (STC) and spatial-temporal matching (STM).
In addition, we use the established masked language modeling (MLM), vision-text\footnote{Vision can either be video or image depending on the dataset.}  contrastive learning (VTC), and vision-text matching (VTM) similar to \cite{vindlu,blip2,li2021align}.

\paragraph{Spatial-Temporal Contrastive Learning} aims to better align the spatial and temporal features of video data.
To this end, we use output features of the last multi-modal exert layer\footnote{Index dropped for clarity.} and learn a cosine similarity function 
\begin{equation}
    \mathrm{s}(\mathbf{V}^{\textrm{spa}},\mathbf{V}^{\textrm{tmp}}) = \Theta_\textrm{spa}(\mathbf{V}^{\textrm{spa}})^\top \Theta_\textrm{tmp}(\mathbf{V}^{\textrm{tmp}}),
\end{equation}
so that aligned spatial-temporal features result in higher similarity scores, where $\Theta_*$ are linear layers that map the features to a normalized lower dimensional vector space.
Then, given spatial and temporal feature pairs, we compute the softmax normalized spatial-to-temporal and temporal-to-spatial similarities as 
\begin{small}  
\begin{align}
    p^\textrm{s2t}_i(\mathbf{V}^\textrm{spa}) &= \frac{\textrm{exp}(\tilde{\mathrm{s}}(\mathbf{V}^\textrm{spa}, \mathbf{V}^\textrm{tmp}_{i}) / \tau)}{\sum_{k=1}^{K} \textrm{exp}(\tilde{\mathrm{s}}(\mathbf{V}^\textrm{spa}, \mathbf{V}^\textrm{tmp}_{k})/\tau)},\label{eq:p_s2t}\\
    p^\textrm{t2s}_i(\mathbf{V}^\textrm{tmp}) &= \frac{\textrm{exp}(\tilde{\mathrm{s}}(\mathbf{V}^\textrm{tmp}, \mathbf{V}^\textrm{spa}_{i}) / \tau)}{\sum_{k=1}^{K} \textrm{exp}(\tilde{\mathrm{s}}(\mathbf{V}^\textrm{tmp}, \mathbf{V}^\textrm{spa}_{k})/\tau)}\label{eq:p_t2s},    
\end{align}
\end{small}
where $\tau$ is learnable temperature parameters, and $\tilde{\textrm{s}}$ is the maximum value of $\mathrm{s}$ as in \cite{blip2}.  
Finally, we can compute the loss as the cross-entropy $\mathcal{H}$ between $\mathbf{p}$ and $\mathbf{y}$:
\begin{small}
\begin{equation}
    \mathcal{L}_{\textrm{stc}} = \frac{1}{2}\mathbb{E}_{(\mathbf{V}^\textrm{spa},\mathbf{V}^\textrm{tmp})}\left[\mathcal{H}\left(\mathbf{y}^\textrm{s2t}, \mathbf{p}^\textrm{s2t}\right) +  \mathcal{H}\left(\mathbf{y}^\textrm{t2s}, \mathbf{p}^\textrm{t2s}\right)\right],
\end{equation}
\end{small}
where $\mathbf{y}^\textrm{s2t}$ and $\mathbf{y}^\textrm{t2s}$ are the golden one-hot similarities.

\begin{table}[!t]
    \centering
    \scalebox{0.7}[0.7]{
        \begin{tabular}{cccc}
        \toprule
         & \textbf{Tr. Stage} \parbox[c]{1.5em}{\includegraphics[width=0.15in]{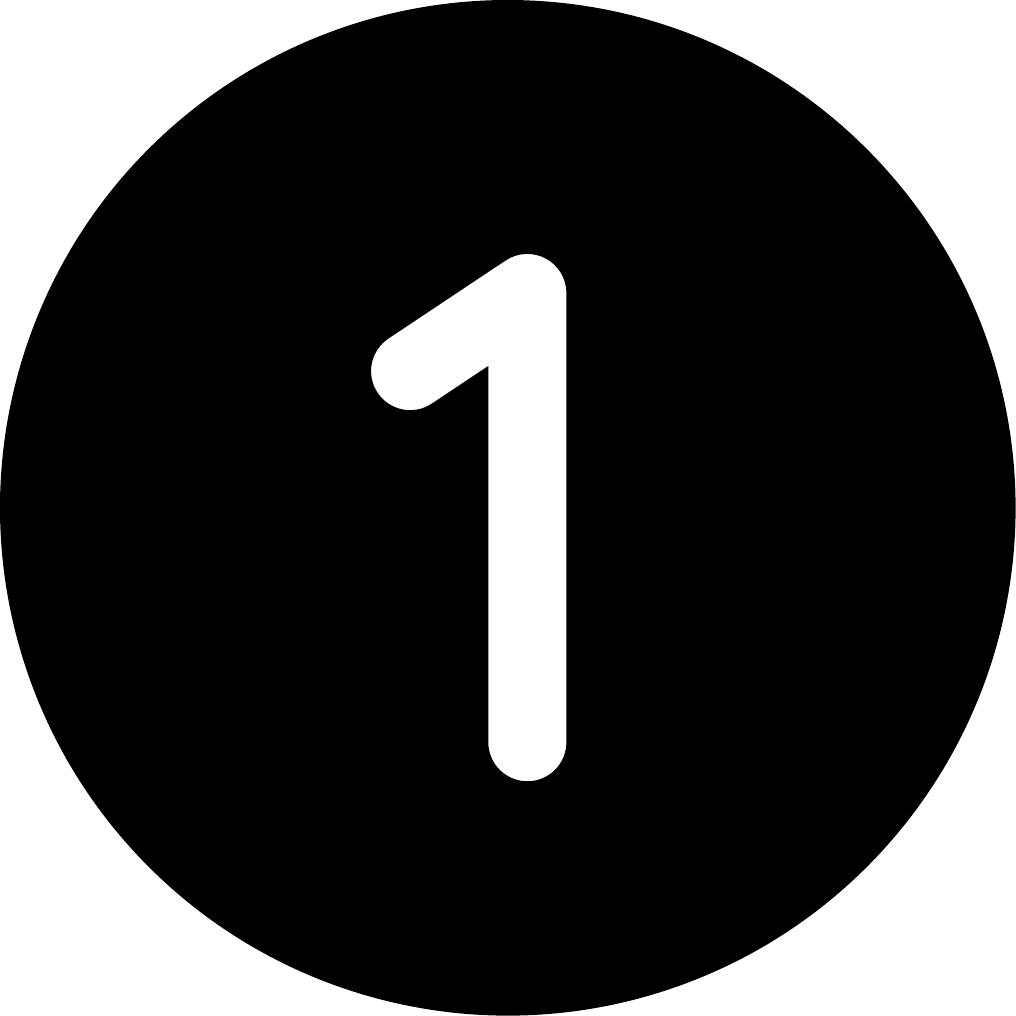}} & \textbf{Tr. Stage} \parbox[c]{1.5em}{\includegraphics[width=0.15in]{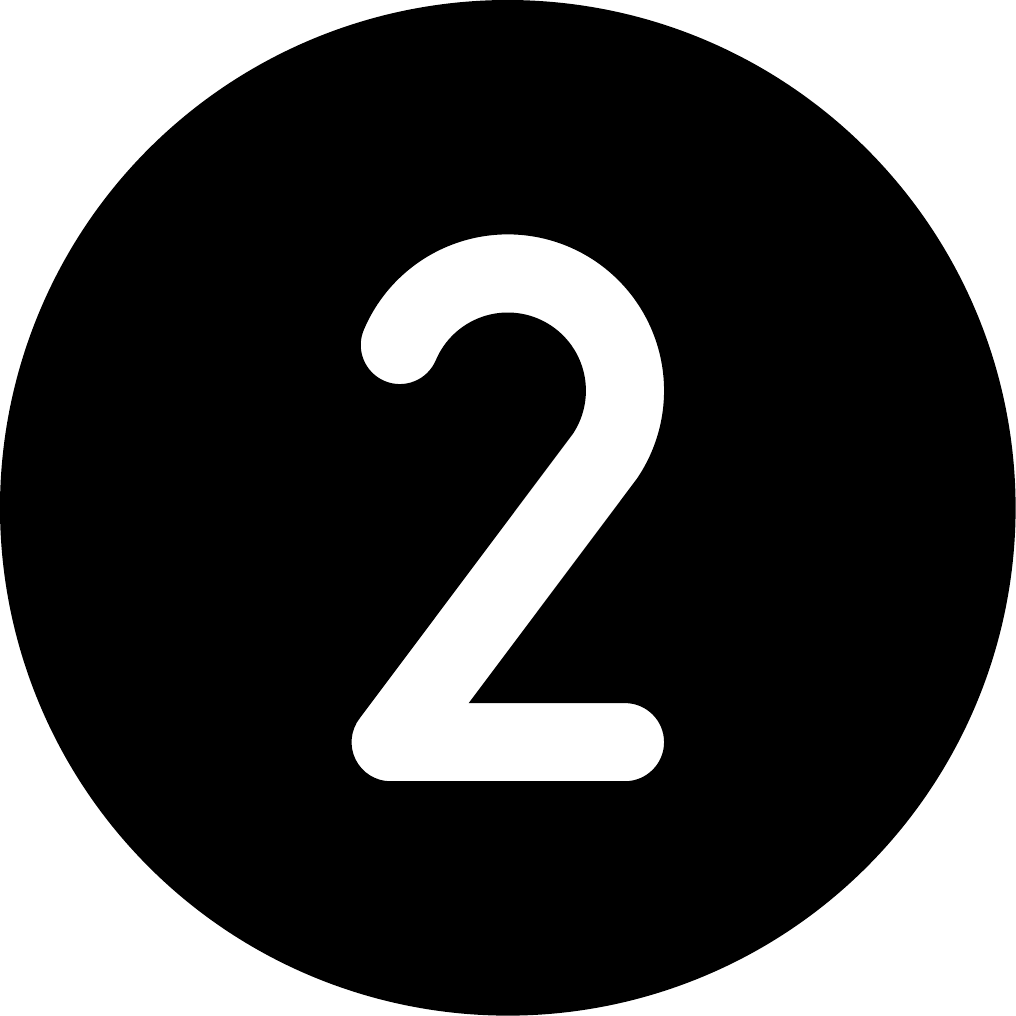}} & \textbf{Tr. Stage} \parbox[c]{1.5em}{\includegraphics[width=0.15in]{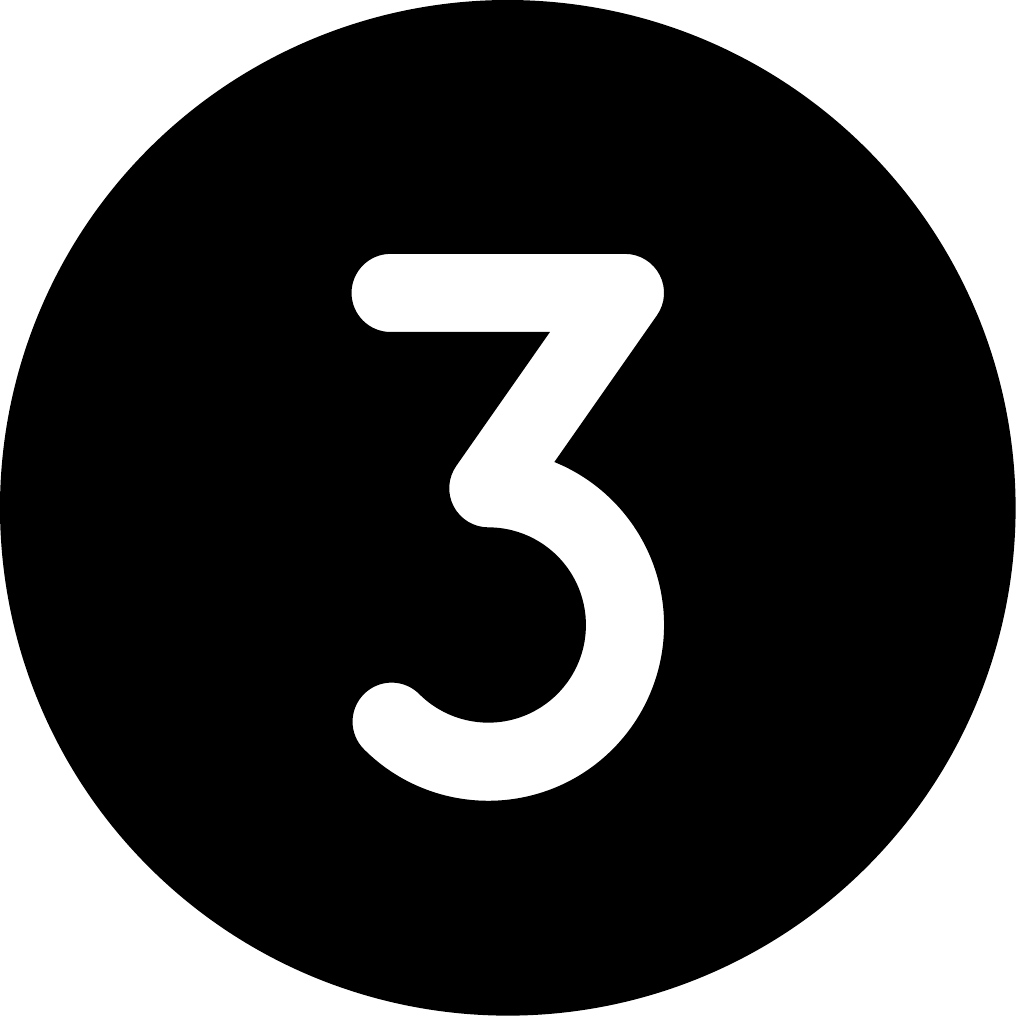}} \\
        \midrule
        \textbf{Videos} \parbox[c]{1.5em}{\includegraphics[width=0.15in]{figures/video_logo.pdf}} &
        $\mathbf{V}^\textrm{spa}, \mathbf{V}^\textrm{tmp}, \mathbf{T}^\textrm{cap}$ & 
        $\mathbf{V}^\textrm{spa}, \mathbf{V}^\textrm{tmp}, \mathbf{T}^\textrm{cap}, \mathbf{T}^\textrm{ctx}$ &
        $\mathbf{V}^\textrm{spa}, \mathbf{V}^\textrm{tmp}, \mathbf{T}^\textrm{cap}, \mathbf{T}^\textrm{ctx}$ \\
        \textbf{Images} \parbox[c]{1.5em}{\includegraphics[width=0.15in]{figures/image_logo.pdf}} &
        $\mathbf{V}^\textrm{spa}, \mathbf{T}^\textrm{cap}$ & - &
        $\mathbf{V}^\textrm{spa}, \mathbf{T}^\textrm{cap}, \mathbf{T}^\textrm{ctx}$ \\
        \bottomrule                                            
\end{tabular}
    }
\caption{Overview of the available features for each training stage and visual input type.}
    \label{tab:input_to_expert_layers}
\end{table}

\paragraph{Spatial-Temporal Matching} complements STC and teaches the model to distinguish between positive and negative spatial-temporal feature pairs. Specifically, a matched feature pair originates from the same video, whereas an unmatched pair is constructed using negative sampling from a different video.
We use a classification token as a proxy of the joint spatial-temporal representations to learn a binary classification problem using the STM loss
\begin{equation}
    \mathcal{L}_\textrm{stm} = \mathbb{E}_{(\mathbf{V}^\textrm{spa}, \mathbf{V}^\textrm{tmp})}\left[\mathcal{H}(\mathbf{y}^\textrm{stm}, \mathbf{p}^\textrm{stm})\right],
\end{equation}
where $\mathbf{p}^\textrm{stm}$ and $\mathbf{y}^\textrm{stm}$ are the predicted and the ground-truth two-class probabilities, respectively.

We provide more details about the remaining established objectives (i.e., MLM, VTC, VTM)  in the supplementary. 

\begin{table*}[!t]
    \centering
    \scalebox{0.62}[0.62]{
        
        \begin{tabular}{l ccccccc  ccccccc ccccccc}
        \toprule
        \multirow{2}*{\textbf{\,\,\,\,Model}} & \multicolumn{7}{c}{\textbf{AVSD-DSTC10}} & \multicolumn{7}{c}{\textbf{AVSD-DSTC8}}  & \multicolumn{7}{c}{\textbf{AVSD-DSTC7}}\\
        \cmidrule(r){2-8} \cmidrule(r){9-15} \cmidrule(r){16-22}
        &\textbf{B-1} & \textbf{B-2} & \textbf{B-3} & \textbf{B-4} & \textbf{M} & \textbf{R} & \textbf{C}
        &\textbf{B-1} & \textbf{B-2} & \textbf{B-3} & \textbf{B-4} & \textbf{M} & \textbf{R} & \textbf{C}
        &\textbf{B-1} & \textbf{B-2} & \textbf{B-3} & \textbf{B-4} & \textbf{M} & \textbf{R} & \textbf{C} \\
        \midrule

$^\vardiamondsuit$MoE-LLAVA$_\textit{\textcolor{gray}{arXiv'24}}$ \cite{moe_llava}    
& $35.8$ & $18.9$ & $10.1$ & $5.9$ & $15.4$ & $27.1$ & $12.8$
& {$39.8$} & $23.9$& $15.2$ & $10.1$ & $18.7$ & $32.2$ & $23.7$
& {$44.7$} & $29.1$ & $19.6$ & $13.8$ & \underline{$21.8$} & $37.3$ & $33.2$ \\

$^\vardiamondsuit$MiniGPT4-video$_\textit{\textcolor{gray}{CVPR'24}}$ \cite{minigpt4_video} 
& {$37.9$} & $19.9$ & $11.3$ & $6.8$ & $16.2$ & $28.7$ & $17.7$ 
& $34.8$ & $17.6$ & $9.7$ & $5.8$ & $15.8$ & $26.3$ & $13.3$
& $37.8$ & $21.2$ & $12.7$ & $8.2$ & $18.4$ & $30.2$ & $17.7$ 
\\

$^\vardiamondsuit$Video-ChatGPT$_\textit{\textcolor{gray}{ACL'24}}$ \cite{video_chatgpt}
& $24.5$ & $14.7$ & $8.8$ & $5.4$ & $16.7$ & $25.2$ & $3.9$ 
& $25.5$ & $16.0$ & $10.1$ & $6.4$ & $18.4$ & $27.1$ & $9.1$
& $28.5$ & $18.5$ & $11.8$ & $7.6$ & $20.4$ & $32.1$ & $19.1$ 
\\

$^\vardiamondsuit$MST-MIXER$_\textit{\textcolor{gray}{ECCV'24}}$ \cite{mst_mixer} 
& $0.1$ & $0.0$ & $0.0$ & $0.0$ & $3.1$ & $6.8$ & $3.0$
& $0.2$ & $0.1$ & $0.1$ & $0.0$ & $3.3$ & $7.1$ & $4.3$
& $0.2$ & $0.1$ & $0.0$ & $0.0$ & $3.4$ & $6.9$ & $4.6$ 
\\

$^\vardiamondsuit$VideoChat2$_\textit{\textcolor{gray}{CVPR'24}}$ \cite{videochat2} 
& \underline{$42.5$} & \underline{$25.9$} & \underline{$16.0$} & \underline{$10.3$} & $\underline{18.7}$ & \underline{$33.1$} & \underline{$25.4$} 
& \underline{$43.9$} & \underline{$28.1$} & $\underline{18.5}$ & \underline{$12.6$} & $\mathbf{20.8}$ & $\underline{34.5}$ & $\underline{29.2}$ 
& \underline{$46.7$} & \underline{$31.1$} & \underline{$20.9$} & \underline{$14.4$} & $\mathbf{22.9}$ & $\underline{37.6}$ & $\underline{31.4}$ 
\\
\midrule
\rowcolor{zscolor}  \,\,\,\,\modelname  \parbox[c]{1.5em}{\includegraphics[width=0.15in]{figures/robot_mixed.png}}
& $\mathbf{54.6}$ & $\mathbf{34.8}$ & $\mathbf{24.0}$ & $\mathbf{17.2}$ & $\mathbf{19.7}$ & $\mathbf{38.3}$ & $\mathbf{53.8}$
& $\mathbf{53.2}$ & $\mathbf{33.8}$ & $\mathbf{23.5}$ & $\mathbf{16.7}$ & $\underline{18.8}$ & $\mathbf{37.7}$ & $\mathbf{49.7}$
& $\mathbf{55.5}$ & $\mathbf{36.7}$ & $\mathbf{26.2}$ & $\mathbf{18.7}$ & ${20.0}$ & $\mathbf{39.2}$ & $\mathbf{50.8}$ \\
\bottomrule                                            
\end{tabular}
    }
\caption{Zero-shot performance comparison on  AVSD-DSTC10, AVSD-DSTC8 and AVSD-DSTC7. Best and second-best performances are in \textbf{bold} and \underline{underlined}.
$\vardiamondsuit$ indicates that we evaluated the model.
(\textbf{B-n} = BLEU-n, \textbf{M} = METEOR, \textbf{R} = ROUGE-L, \textbf{C} = CIDEr).
}
\label{tab:avsd_zeroshot}
\end{table*}

\subsubsection{Stages 2 \& 3}
In the subsequent stages, we couple the multimodal expert layers with a pre-trained Flan-T5$_\texttt{large}$ \cite{flant5} via a linear layer.
Specifically, Stage 2 aims to align the hidden states of the proposed layers with those of the pre-trained LLM.
To this end, we keep the LLM weights frozen and train the whole architecture end-to-end using the generative loss (i.e., next token prediction) on large scale video dialog data
% \footnote{The weights of $\mathcal{E}_\textrm{ctx}$ are initialized with those of $\mathcal{E}_\textrm{cap}$ from Stage 1.}
, i.e., 
\begin{align}
    \mathcal{L}_\textrm{gen} &= \mathbb{E}_{\mathbf{X}^\mathrm{gen}} \left[\mathcal{H}(\mathbf{y}^\textrm{gen}_\rightarrow, \mathbf{p}^\textrm{gen})\right],\\
    \mathbf{X}^\mathrm{gen} &= \Theta_\textrm{gen}\left( \mathrm{LLM}_\textrm{dec}([\mathbf{X}^\textrm{enc}, \mathbf{T}^\textrm{ans}])\right),
\end{align}
where $\mathbf{X}^\textrm{enc}$, $\mathbf{T}^\textrm{ans}$ and $\Theta_\textrm{gen}$ are the LLM encoder output, the answer token embeddings, and a linear layer that maps the features to the vocabulary space, respectively.
$\mathbf{y}^\textrm{gen}_\rightarrow$ and $\mathbf{p}^\textrm{gen}$ denote the right-shifted ground-truth answer tokens and the predicted text token probabilities.
Finally, in Stage 3, we unfreeze the LLM weights and fine-tune our model end-to-end on the downstream tasks of video and visual dialog using the same generative loss.

\section{Experiments}
\subsection{Datasets}
As shown in \autoref{fig:pipeline}, we simultaneously use the video and image captioning datasets of WebVid-2M \cite{webvid} and CC-3M \cite{cc3m}  to pre-train the multimodal expert layers in Stage 1. 
Then in the \textit{second} stage, we use 25\% of the recent large-scale video dialog dataset Champagne \cite{champagne} before performing \textit{zero-shot} evaluation on the widely used video and visual dialog datasets of AVSD \cite{avsd} and VisDial \cite{visdial}, respectively. Finally, in the \textit{third} stage, we perform a domain shift evaluation based on different combinations of AVSD and VisDial to quantify whether and to what extent these seemingly similar benchmarks benefit from each other in both \textit{zero-shot} and \textit{fine-tuning} evaluation settings.

\subsection{Evaluation Metrics}
\begin{figure}[!t]
    \begin{minipage}{1\linewidth}
        \centering
        \scalebox{0.98}[0.98]{
            \includegraphics[width=\textwidth]{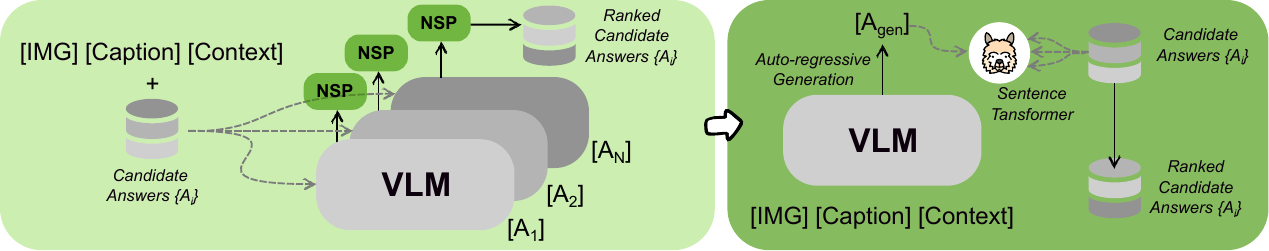}
        }
        \caption{Instead of training a dedicated NSP head, we propose a ranking scheme based on the cosine similarity of the candidate answers' embeddings with the respect to those of the generated ones.
        We used RoBERTa$_\texttt{large}$ \cite{roberta} and OpenAI Text Embedding-3$_\texttt{large}$ to generate these embeddings.
        }
        \label{fig:novel_visdial_eval}
    \end{minipage}
\end{figure}

We use the established official metrics for each dataset to fairly benchmark \modelname\, with previous works.
Specifically, for all \textit{three} AVSD datasets, we use BLEU (\textbf{B-n}) \cite{bleu}, ROUGE-L (\textbf{R}) \cite{rouge}, METEOR (\textbf{M}) \cite{meteor}, and CIDEr (\textbf{C}) \cite{cider}.
Whereas for VisDial, we use the retrieval metrics of recall (\textbf{R@k}), mean reciprocal rank (\textbf{MRR}), and normalized discounted cumulative gain (\textbf{NDCG}).
However, since we are jointly tackling both tasks with a fully generative model, we propose to rank the VisDial candidate answers by means of cosine similarity with respect to the generated answer using the embdeddings of a pre-trained sentence transformer (i.e. RoBERTa \cite{roberta} and OpenAI Text Embedding-3).
We posit that this approach is more natural, caters to the current advances in generative models, and appropriately captures the semantic similarities between the generated and the candidate answers.
In addition, it allows for a seamless unification of AVSD and VisDial without any training or architectural modifications.
As shown in \autoref{fig:novel_visdial_eval}, our proposed adaptation does \textit{not} alter the computation of the sparse metrics itself and \textit{only} \textit{rethinks} the ranking of the candidate answers allowing for a fair comparison with previous works.

\subsection{Experimental Setup}
In the first stage, we trained our model for a maximum of \textit{ten} epochs and applied early stopping based on a validation split to select the best checkpoint. 
In the subsequent stages, we trained it for up to \textit{three} and \textit{twelve} epochs, respectively.
In all stages, we used the AdamW \cite{adamw} optimizer with the default parameters and a weight decay value of $0.01$.
Furthermore, we applied a linear learning rate schedule with warm-up and minimum and base values of $5e-5$ and $1e-4$, respectively.
We conducted our experiments on a cluster consisting of \textit{eight} A100 GPUs. 

\subsection{Zero-shot Evaluation}
\paragraph{AVSD.}
We first assessed \modelname in a zero-shot setting on AVSD.
This is in stark contrast to previous models that were exclusively evaluated in a fine-tuning setting.
We instead advocate for complementing the fine-tuning evaluation setting with a zero-shot one, as it results in a more rigorous and challenging testbed for the proposed models.
To this end, we establish; to the best of our knowledge; the \textit{first} zero-shot benchmark comparison on AVSD comprised of recent capable multimodal models.
As can be seen from \autoref{tab:avsd_zeroshot}, our model outperforms all baselines by a considerable margin across 6/7 metrics of AVSD-DSTC8 and AVSD-DSTC7.
On the more recent and challenging version of the benchmark (i.e. AVSD-DSTC10), \modelname ranks first across all metrics.
For instance, it more than doubles the CIDEr score compared to VideoChat2 \cite{videochat2}.

 \begin{table}[!t]
    \centering
    \scalebox{0.63}[0.63]{
          \begin{tabular}{lcccccc}
            \toprule
            \textbf{\,\,\,\,Model} & \textbf{Sent. Trans.} & \textbf{R@1} & \textbf{R@5} & \textbf{R@10} & \textbf{MRR}
            & \textbf{NDCG} \\
            \midrule
            \,\,\,\,FROMAGe$_\textit{\textcolor{gray}{ICML'23}}$   \cite{koh2023grounding} & & ${17.6}$ & ${20.1}$ &${25.1}$  & ${22.0}$ & $16.5$  \\
            
            \,\,\,\,ESPER$_\textit{\textcolor{gray}{CVPR'23}}$     \cite{yu2023fusing} & \textit{n/a} & $14.6$ & $-$ & $-$  &  ${25.7}$ & ${22.3}$  \\

            \,\,\,\,Champagne$_\textit{\textcolor{gray}{ICCV'23}}$ \cite{champagne}  & & $-$ &$-$ &$-$ &$-$ &${25.5}$ \\
            
            \hdashline
            
            $^\vardiamondsuit$MoE-LLAVA$_\textit{\textcolor{gray}{arXiv'24}}$ \cite{moe_llava}  &  & $10.6$ &$25.4$ &$36.4$ &$19.6$ &$26.7$ \\
            
            $^\vardiamondsuit$MiniGPT-video$_\textit{\textcolor{gray}{CVPR'24}}$ \cite{minigpt4_video}  &  & $7.4$ &$17.4$ &$26.5$ &$14.6$ &$23.2$ \\
            
            $^\vardiamondsuit$Video-ChatGPT$_\textit{\textcolor{gray}{ACL'24}}$ \cite{video_chatgpt}  & \textit{RoBERTa} & $10.0$ & $22.5$ & $31.5$ & $18.1$ &$24.8$ \\
            
            $^\vardiamondsuit$MST-MIXER$_\textit{\textcolor{gray}{ECCV'24}}$ \cite{mst_mixer}  &  & {$18.2$} &$22.1$ &$25.7$ &$21.9$ &$24.6$ \\
            
            $^\vardiamondsuit$VideoChat2$_\textit{\textcolor{gray}{CVPR'24}}$ \cite{videochat2}  &  & $12.7$ & {$29.0$} & $\underline{39.9}$ & $22.3$ & $30.9$\\
            
            \midrule
            \rowcolor{zscolor}  & \textit{RoBERTa} & $\underline{20.0}$ & $\underline{30.2}$ & ${39.3}$  & $\underline{26.9}$ & $\mathbf{33.3}$ \\
            
            \rowcolor{zscolor}\multirow{-2}*{\,\, \modelname \parbox[c]{1.5em}{\includegraphics[width=0.15in]{figures/robot_mixed.png}}} & \textit{OpenAI TE-3} & $\mathbf{22.1}$ & $\mathbf{41.2}$ & $\mathbf{48.1}$  & $\mathbf{32.7}$ & $\underline{32.0}$ \\
            \bottomrule
          \end{tabular}
          }
        \caption{Zero-shot performance comparison on the VisDial v1.0 val split. OpenAI TE-3 = OpenAI Text Embedding-3$_\texttt{large}$.
      }
    \label{tab:visdial_zeroshot}
\end{table}

\begin{figure*}[!t]
    \begin{minipage}{1\linewidth}
        \centering
        \scalebox{0.9}[0.9]{
            \includegraphics[width=\textwidth]{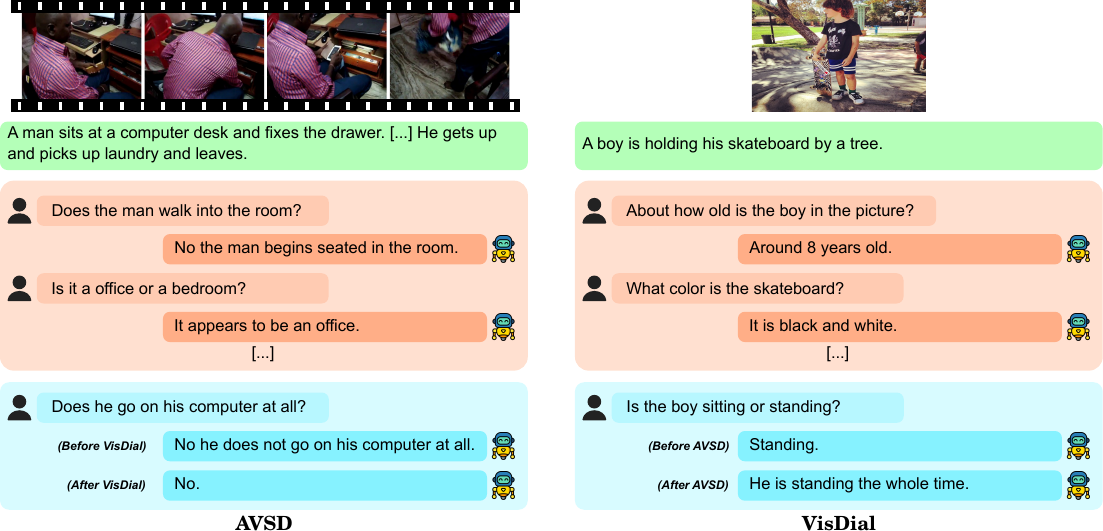}
        }
        \caption{Zero-shot qualitative examples of \modelname before and after fine-tuning on VisDial and AVSD. The former teaches the model to answer question with brief responses whereas the latter teaches it to produce longer and more elaborate answers.}
        \label{fig:qualitative}
    \end{minipage}
\end{figure*}

\begin{table*}[!t]
    
    \centering
    \scalebox{0.6}[0.6]{
        
        \begin{tabular}{l ccccccccccccccccccccc}
        \toprule
        \multirow{2}*{\textbf{\,\,\,\,Model}} & \multicolumn{7}{c}{\textbf{AVSD-DSTC10}} & \multicolumn{7}{c}{\textbf{AVSD-DSTC8}}  & \multicolumn{7}{c}{\textbf{AVSD-DSTC7}}\\
        \cmidrule(r){2-8} \cmidrule(r){9-15} \cmidrule(r){16-22}
        & \textbf{B-1} & \textbf{B-2} & \textbf{B-3} & \textbf{B-4} & \textbf{M} & \textbf{R} & \textbf{C}
        &\textbf{B-1} & \textbf{B-2} & \textbf{B-3} & \textbf{B-4} & \textbf{M} & \textbf{R} & \textbf{C}
        &\textbf{B-1} & \textbf{B-2} & \textbf{B-3} & \textbf{B-4} & \textbf{M} & \textbf{R} & \textbf{C} \\
        \midrule

\,\,\,\,PDC$_\textit{\textcolor{gray}{ICLR'21}}$ \cite{le2021learning}    
& $-$ & $-$ & $-$ & $-$ & $-$ & $-$ & $-$ 
& $74.9$ & $62.9$ & $52.8$ & $43.9$ & $28.5$ & $59.2$ & $120.1$
& $77.0$ & $65.3$ & $53.9$ & $44.9$ & $29.2$ & $60.6$ & $129.5$
\\

\,\,\,\,THAM$_\textit{\textcolor{gray}{EMNLP'22}}$ \cite{tham} 
& $-$ & $-$ & $-$ & $-$ & $-$ & $-$ & $-$ 
& $76.4$ & $64.1$ & $53.8$ & $45.5$ & $30.1$ & $61.0$ & $130.4$
& $77.8$ & $65.4$ & $54.9$ & $46.8$ & $30.8$ & $61.9$ & $133.5$ 
\\

\,\,\,\,DialogMCF$_\textit{\textcolor{gray}{TASLP'23}}$ \cite{dmcf} 
& $69.3$ & $55.6$ & $45.0$ & $36.9$ & $24.9$ & $53.6$ & $91.2$
& $75.6$ & $63.3$ & $53.2$ & $44.9$ & $29.3$ & $60.1$ & $125.3$
& $77.7$ & $65.3$ & $54.7$ & $45.7$ & $30.6$ & $61.3$ & $135.2$
\\

$^\vardiamondsuit$VideoLLAMA 2$_\textit{\textcolor{gray}{arXiv'24}}$ \cite{video_llama2} 
& $50.2$ & $35.0$ & $24.9$ & $18.1$ & $21.8$ & $42.8$ & $57.5$
& $53.3$ & $39.0$ & $29.1$ & $22.2$ & $24.8$ & $46.3$ & $74.0$
& $56.2$ & $41.1$ & $30.7$ & $23.2$ & $26.4$ & $48.5$ & $79.2$
\\

\,\,\,\,MST-MIXER$_\textit{\textcolor{gray}{ECCV'24}}$ \cite{mst_mixer} 
&$\underline{69.7}$     & $\underline{57.1}$     & $\underline{47.2}$     & $\underline{39.5}$     & $\underline{25.1}$     & $\underline{54.0}$     & $\underline{96.9}$
& $\mathbf{77.1}$     & $\mathbf{65.6}$     & $\underline{55.7}$     & $\underline{47.1}$   & $\underline{30.2}$     & $\underline{61.8}$     & $\underline{133.6}$
&$\underline{78.4}$    & $\underline{66.0}$     & $\underline{55.8}$     & $\underline{47.1}$     & $\underline{31.0}$     & $\underline{62.0}$     & $\underline{136.5}$
\\

\midrule
\rowcolor{ftcolor}  \,\,\,\,\modelname  \parbox[c]{1.5em}{\includegraphics[width=0.15in]{figures/robot_mixed.png}} & 
$\mathbf{70.7}$     & $\mathbf{58.2}$     & $\mathbf{48.2}$     & $\mathbf{40.3}$     & $\mathbf{26.0}$     & $\mathbf{55.4}$     & $\mathbf{103.3}$

& $\underline{76.8}$     & $\underline{65.5}$     & $\mathbf{55.8}$     & $\mathbf{47.5}$   & $\mathbf{30.4}$     & $\mathbf{62.1}$     & $\mathbf{135.7}$

&$\mathbf{78.9}$    & $\mathbf{66.5}$     & $\mathbf{56.1}$     & $\mathbf{47.4}$     & $\mathbf{31.2}$     & $\mathbf{62.3}$     & $\mathbf{139.8}$

\\

\bottomrule                                            
\end{tabular}
    }
\caption{Fine-tuning performance comparison on  AVSD-DSTC10, AVSD-DSTC8 and AVSD-DSTC7. 
VideoLLAMA 2 \cite{video_llama2} was trained on AVSD amongst other datasets.
Additional model comparisons can be found in the supplementary material.
}
\label{tab:avsd_finetuned}
\end{table*}

\begin{table}[!t]
    \centering
        \scalebox{0.65}[0.65]{
          \begin{tabular}{lcccccc}
            \toprule
            \textbf{\,\,\,\,Model} & \textbf{Sent. Trans.} &  \textbf{R@1}& \textbf{R@5} & \textbf{R@10} & \textbf{MRR}  & \textbf{NDCG} \\
            \midrule
            \,\,\,\,{LTMI}$_\textit{\textcolor{gray}{ECCV'20}}$ \cite{Nguyen2020} &  & $40.4$ & $\underline{61.6}$ & $\mathbf{69.7}$& $50.7$  & $\mathbf{63.5}$  \\
            \,\,\,\,{LTMI-LG}$_\textit{\textcolor{gray}{EMNLP'21}}$  \cite{ltmi_lg} & &   $41.3$ & $\underline{61.6}$ & $69.0$ & $\underline{51.3}$ & $\underline{63.2}$  \\
            \,\,\,\,{GoG}$_\textit{\textcolor{gray}{ACL'21}}$  \cite{GOG}             & \textit{n/a} &  $41.2$ & $\mathbf{61.8}$ & $\underline{69.4}$& $\underline{51.3}$ & $62.6$   \\
            \,\,\,\,{UTC}$_\textit{\textcolor{gray}{CVPR'22}}$  \cite{Chen2022}            & & $41.3$ & $59.8$ & $66.3$ & $50.6$ & $61.0$ \\
            \,\,\,\,{Champagne}$_\textit{\textcolor{gray}{ICCV'23}}$  \cite{champagne}  & &  $-$ & $-$ & $-$ & $-$ & $62.5$ \\
            \hdashline
            $^\clubsuit${MST-MIXER}$_\textit{\textcolor{gray}{ECCV'24}}$ \cite{mst_mixer}     & \textit{RoBERTa} &  $42.2$ & $51.6$ & $57.8$ & $47.7$ & $52.5$ \\
            \midrule
            \rowcolor{ftcolor} & \textit{RoBERTa} & $\underline{45.4}$ & $54.7$ & $61.1$ & $50.9$ & $54.0$ \\
            
            \rowcolor{ftcolor} \multirow{-2}*{\,\,\,\,\modelname \parbox[c]{1.5em}{\includegraphics[width=0.15in]{figures/robot_mixed.png}}} & \textit{OpenAI TE-3} & $\mathbf{46.1}$ & $59.3$ & $65.7$ & $\mathbf{53.2}$ & $53.1$ \\
            \bottomrule
          \end{tabular}
          }
        \caption{Fine-tuning performance comparison on the VisDial v1.0 val split. $\clubsuit$ indicates that we trained and evaluated the model. 
      }
    \label{tab:visdial_finetuned}
\end{table}

\paragraph{VisDial.}
Additionally, we assessed the same model checkpoint on VisDial v1.0.
As can be seen from \autoref{tab:visdial_zeroshot}, \modelname managed to outperform previous models such as FROMAGe \cite{koh2023grounding} by a considerable margin across all metrics of the dataset.
In addition, it outperformed Champagne \cite{champagne} that was trained on x4 more dialog data by $7.8$ absolute NDCG points. 
Furthermore, our model outperformed the more recent baselines of the previous section on 4/5 metrics, underlining it capability of dealing with both video and image input data types.
Finally, replacing the sentence embeddings generated by RoBERTa$_\texttt{large}$ with those from OpenAI Text Emedding-3 improved the external ranking of the candidate answers and resulted in higher scores across all metrics, as can be seen in the last row of \autoref{tab:visdial_zeroshot}. 

\subsection{Fine-tuning Evaluation}
\paragraph{AVSD.}
Similar to almost all previous works on AVSD, we assessed \modelname in a fine-tuning setting on all \textit{three} benchmarks of the dataset.
As can be seen from \autoref{tab:avsd_finetuned}, our model managed to maintain it competitiveness ahead of recent models and outperformed them on the latest and most challenging AVSD-DSTC10 benchmarks across all evaluation metrics.
For instance, it lifted CIDEr by over $6$ absolute points compared to the second-best model.
Furthermore, our model managed to maintain an on par performance with the state of the art on AVSD-DSTC8 and AVSD-DSTC7.
As shown in \autoref{tab:avsd_finetuned}, \modelname\, increased their respective CIDEr scores by over $2$ and $3$ absolute points compared to the second-best model.
\paragraph{VisDial.}
Finally, we fine-tuned our model and MST-MIXER \cite{mst_mixer} that had the closest AVSD performance on Visdial v1.0 using the same fully-generative approach.
As can be seen from \autoref{tab:visdial_finetuned}, \modelname managed to outperform all previous models on the strictest metric of the dataset by achieving a R@1 score of $44.2$.
However, when using OpenAI Text Embedding-3 our model managed to increase the R@1 and MRR scores to $44.9$ and $52.4$, respectively, thereby setting new state-of-the-art results. 
As expected and due to the more challenging aspect of a tackling VisDial as a fully generative task, our model performed slightly worse than the previous fine-tuned models on the remaining metrics of the dataset.
However, when comparing our model with MST-MIXER that was trained using the same paradigm (i.e. the last two rows of \autoref{tab:visdial_finetuned}), we can see that our model outperformed it across 4/5 metrics of the task and scored almost equally on NDCG.

\subsection{Domain Shift Evaluation}
\paragraph{Zero-shot setting.}
First, we fine-tuned our model's checkpoint from Stage 2 on AVSD and zero-shot evaluated it on VisDial.
As can be seen from the second section of \autoref{tab:domain_shift_evaluation}, our model's performance was lifted by a considerable margin across most metrics.
Notably, the NDCG score improved by $9$ absolute points compared to the results of \autoref{tab:visdial_zeroshot}.
Then, we replicated the same experiment on AVSD after having fine-tuned the model on VisDial.
Interestingly, our model's performance deteriorated across all metrics of the benchmark.
This behavior could be explained by the nature of both datasets.
Whereas AVSD encourages the model to produce long and  elaborate responses, VisDial teaches it to produce brief answers instead, which diminishes its performance on the language generation metrics.
The qualitative examples of \autoref{fig:qualitative} clearly illustrate this phenomenon on both datasets.

\paragraph{Fine-tuning Setting.}
We first experimented with a curriculum learning strategy where we used one dataset for pre-training before finally fine-tuning on the other.
As can be seen from the last section of \autoref{tab:domain_shift_evaluation}, this training paradigm resulted in performance drops on both datasets compared to \autoref{tab:avsd_finetuned} and \autoref{tab:visdial_finetuned} where the model was only trained on the data of the respective benchmark.
This indicates that the converged model's weights on one dataset do not offer a good  initialization for training on the remaining one. 
Allowed by our model design that can jointly handle video and image input data, we finally fine-tuned one single model on both datasets simultaneously.
\noindent As seen from the last row of \autoref{tab:domain_shift_evaluation}, this resulted in the best joint performance of our model across the two datasets.
Although the results on AVSD slightly dropped compared to \autoref{tab:avsd_finetuned}, our model lifted its performance on VisDial by a considerable margin.
This could largely be attributed to the same previous observation, as training on VisDial incentivizes our model to shorten its responses on AVSD.
Additional qualitative examples can be found in the supplementary material.
\begin{table*}[!t]
    
    \centering
    \scalebox{0.7}[0.7]{
        
        \begin{tabular}{ccc cccc cccc cccc cccc}
        \toprule
        \multicolumn{3}{c}{\textbf{Fine-tuning data}} & \multicolumn{4}{c}{\textbf{AVSD-DSTC10}} & \multicolumn{4}{c}{\textbf{AVSD-DSTC8}} & \multicolumn{4}{c}{\textbf{AVSD-DSTC7}} & \multicolumn{4}{c}{\textbf{VisDial}} \\
        
        \cmidrule(r){1-3} \cmidrule(r){4-7} \cmidrule(r){8-11} \cmidrule(r){12-15} \cmidrule(r){16-19}
        \textbf{AVSD} & & \textbf{VisDial}
        & \textbf{B-1} & \textbf{M} & \textbf{R} & \textbf{C}
        & \textbf{B-1} & \textbf{M} & \textbf{R} & \textbf{C}
        & \textbf{B-1} & \textbf{M} & \textbf{R} & \textbf{C}
        & \textbf{R@1} & \textbf{R@5} & \textbf{R@10} & \textbf{NDCG} \\
        \midrule
         \multirow{2}*{\xmark} & & \multirow{2}*{\xmark} &  \multicolumn{16}{c}{ \cellcolor{zscolor} \textit{Zero-shot (from \autoref{tab:avsd_zeroshot} and \autoref{tab:visdial_zeroshot})}} \\
        & & 
        & \cellcolor{zscolor} $54.6$ & \cellcolor{zscolor} $19.7$ & \cellcolor{zscolor} $38.3$ & \cellcolor{zscolor} $53.8$
        & \cellcolor{zscolor} $53.2$ & \cellcolor{zscolor} $18.8$ & \cellcolor{zscolor} $37.7$ & \cellcolor{zscolor} $49.7$
        & \cellcolor{zscolor} $55.5$ & \cellcolor{zscolor} $20.0$ & \cellcolor{zscolor} $39.2$ & \cellcolor{zscolor} $50.8$
        & \cellcolor{zscolor} $20.0$ & \cellcolor{zscolor} $30.2$ & \cellcolor{zscolor} $39.3$ & \cellcolor{zscolor} $33.3$ \\
        \hdashline
         \multirow{2}*{\cmark} & & \multirow{2}*{\xmark} & \multicolumn{12}{c}{ \cellcolor{ftcolor} \textit{Fine-tuning (from \autoref{tab:avsd_finetuned})}} & \multicolumn{4}{c}{ \cellcolor{zscolor} \textit{Zero-shot}} \\
        & &
        & \cellcolor{ftcolor} $70.7$ & \cellcolor{ftcolor} $26.0$ & \cellcolor{ftcolor} $55.4$ & \cellcolor{ftcolor} $103.3$
        & \cellcolor{ftcolor} $76.8$ & \cellcolor{ftcolor} $30.4$ & \cellcolor{ftcolor} $62.1$ & \cellcolor{ftcolor} $135.7$
        & \cellcolor{ftcolor} $78.9$ & \cellcolor{ftcolor} $31.2$ & \cellcolor{ftcolor} $62.3$ & \cellcolor{ftcolor} $139.8$
        & \cellcolor{zscolor} $12.8$ & \cellcolor{zscolor} $36.7$ & \cellcolor{zscolor} $50.8$ & \cellcolor{zscolor} $42.3$ \\
        \multirow{2}*{\xmark} & & \multirow{2}*{\cmark} & \multicolumn{12}{c}{ \cellcolor{zscolor} \textit{Zero-shot}} & \multicolumn{4}{c}{ \cellcolor{ftcolor} \textit{Fine-tuning (from \autoref{tab:visdial_finetuned})}} \\
        & &  
        & \cellcolor{zscolor} $11.5$  & \cellcolor{zscolor} $6.8$  & \cellcolor{zscolor} $20.1$ & \cellcolor{zscolor} $14.6$
        & \cellcolor{zscolor} $11.5$  & \cellcolor{zscolor} $7.3$  & \cellcolor{zscolor} $20.7$ & \cellcolor{zscolor} $20.9$
        & \cellcolor{zscolor} $7.9$  & \cellcolor{zscolor} $6.2$  & \cellcolor{zscolor} $17.4$ & \cellcolor{zscolor} $18.2$
        & \cellcolor{ftcolor} $44.2$ & \cellcolor{ftcolor} $53.3$ & \cellcolor{ftcolor} $59.5$ & \cellcolor{ftcolor} $52.3$ \\

        \hdashline

        & & & \multicolumn{16}{c}{ \cellcolor{ftcolor} \textit{Fine-tuning}} \\
        \cmark & $\rightarrow$ & \cmark 
        & \cellcolor{ftcolor} $-$ & \cellcolor{ftcolor} $-$ & \cellcolor{ftcolor} $-$ & \cellcolor{ftcolor} $-$        
        & \cellcolor{ftcolor} $-$ & \cellcolor{ftcolor} $-$ & \cellcolor{ftcolor} $-$ & \cellcolor{ftcolor} $-$
        & \cellcolor{ftcolor} $-$ & \cellcolor{ftcolor} $-$ & \cellcolor{ftcolor} $-$ & \cellcolor{ftcolor} $-$
        & \cellcolor{ftcolor} $42.2$ & \cellcolor{ftcolor} $50.1$ & \cellcolor{ftcolor} $56.3$ & \cellcolor{ftcolor} $51.3$ \\

        \cmark & $\leftarrow$ & \cmark 
        & \cellcolor{ftcolor} $69.6$ & \cellcolor{ftcolor} $25.7$ & \cellcolor{ftcolor} $55.0$ & \cellcolor{ftcolor} $100.5$
        & \cellcolor{ftcolor} $75.9$ & \cellcolor{ftcolor} $29.8$ & \cellcolor{ftcolor} $61.4$ & \cellcolor{ftcolor} $132.1$
        & \cellcolor{ftcolor} $77.6$ & \cellcolor{ftcolor} $30.4$ & \cellcolor{ftcolor} $61.5$ & \cellcolor{ftcolor} $134.5$
        & \cellcolor{ftcolor} $-$ & \cellcolor{ftcolor} $-$ & \cellcolor{ftcolor} $-$ & \cellcolor{ftcolor} $-$ \\

        \cmark & $\&$ & \cmark 
        & \cellcolor{ftcolor} $69.3$ & \cellcolor{ftcolor} $25.4$ & \cellcolor{ftcolor} $54.8$ & \cellcolor{ftcolor} $99.9$
        & \cellcolor{ftcolor} $75.1$ & \cellcolor{ftcolor} $29.3$ & \cellcolor{ftcolor} $61.1$ & \cellcolor{ftcolor} $130.0$
        & \cellcolor{ftcolor} $77.3$ & \cellcolor{ftcolor} $30.0$ & \cellcolor{ftcolor} $61.7$ & \cellcolor{ftcolor} $134.5$
        & \cellcolor{ftcolor} $45.4$ & \cellcolor{ftcolor} $54.7$ & \cellcolor{ftcolor} $61.1$ & \cellcolor{ftcolor} $54.0$ \\ 
        
        \bottomrule                                            
\end{tabular}
    }
\caption{
    Domain shift evaluation between the respective \textit{most prominent} video and visual dialog datasets of AVSD and VisDial. $\square \rightarrow \triangle$ means that the model was pre-trained on dataset $\square$ before fine-tuning on dataset $\triangle$. 
        }
    \label{tab:domain_shift_evaluation}
\end{table*}

\begin{table}[!t]
    
    \centering
    \scalebox{0.8}[0.8]{
        
        \begin{tabular}{lcr cccc cc }
        \toprule
        \multicolumn{3}{c}{\textbf{Expert}} & \multicolumn{4}{c}{\textbf{AVSD-DSTC7}} & \multicolumn{2}{c}{\textbf{VisDial}} \\
        
        \cmidrule(r){4-7} \cmidrule(r){8-9}
       \multicolumn{3}{c}{\textbf{Swapping}} 
        & \textbf{B-1} & \textbf{M} & \textbf{R} & \textbf{C}
        & \textbf{R@1} & \textbf{NDCG} \\
        \midrule
         
        \multicolumn{3}{l}{Original}
        & $78.9$ & $31.2$ & $62.3$ & $139.8$
        & $44.2$ & $52.3$ \\
        \hline
        \rowcolor{mygray} \multicolumn{9}{l}{\textit{Swapping experts of the same modality (vision / language)}}\\
        \hdashline
        $\mathcal{E}_\mathrm{spa}$ & $\leftrightarrow$ & $\mathcal{E}_\mathrm{tmp}$
        & $77.0$ & $29.5$ & $61.2$ & $133.7$
        & $-$ & $-$ \\
        $\mathcal{E}_\mathrm{cap}$ & $\leftrightarrow$ & $\mathcal{E}_\mathrm{ctx}$
        & $76.1$ & $29.6$ & $60.3$ & $131.1$&
        $42.8$ & $51.9$
        \\
        \hline
        \rowcolor{mygray}\multicolumn{9}{l}{\textit{Swapping experts of different modalities}}\\
        \hdashline
        $\mathcal{E}_\mathrm{spa}$ & $\leftrightarrow$ & $\mathcal{E}_\mathrm{cap}$ & \multirow{2}*{$28.5$} & \multirow{2}*{$10.7$} & \multirow{2}*{$22.0$} & \multirow{2}*{$10.1$} & $35.7$ & $45.3$\\
        $\mathcal{E}_\mathrm{tmp}$ & $\leftrightarrow$ & $\mathcal{E}_\mathrm{ctx}$ & & & & & $-$ & $-$ \\
        \hdashline
        $\mathcal{E}_\mathrm{spa}$ & $\leftrightarrow$ & $\mathcal{E}_\mathrm{ctx}$ & \multirow{2}*{$34.4$} & \multirow{2}*{$12$} & \multirow{2}*{$25.4$} & \multirow{2}*{$11.7$} & $32.4$ & $42.9$\\
        $\mathcal{E}_\mathrm{tmp}$ & $\leftrightarrow$ & $\mathcal{E}_\mathrm{cap}$  & & & & & $-$ & $-$ \\
        \bottomrule                                            
\end{tabular}
    }
\caption{
    Expert swapping results. $\mathcal{E}_\square \leftrightarrow \mathcal{E}_\triangle$ means that the $\square$ features are \textit{inadequately} routed \textit{at test time} through  $\mathcal{E}_\triangle$ and vice versa.
    The other experts remain unchanged.
    }
    \label{tab:adversarial_attack}
\end{table}
\subsection{Expert Swapping Experiment}
In order to validate the specialization of each expert, we conducted a swapping experiment where we routed some features through \textit{inadequate} experts.
We first swapped experts of the same modality (i.e., experts operating on vision or language data).
As shown in \autoref{tab:adversarial_attack}, this resulted in performance drops across all metrics of both datasets, indicating that experts of the same modality are able to capture the semantic nuances of the data they specialize on.
More interestingly, the performance of our model dropped more significantly when swapping experts of different modalities, as seen from the last section of \autoref{tab:adversarial_attack}.
This showcases their ability to adjust to the nature of the data they process and to capture its modality specific features.
\subsection{Ablation Study}
\paragraph{Effect of Pre-training Data.}
% To assess the effectiveness of the pre-training data in the first two stages,
We trained two versions of our model, where one was only pre-trained on Stage 1 using WebVid-2M \& CC-3M and the other only on Stage 2 with a subset of Champagne.
As can be seen from the middle section of \autoref{tab:ablations}, our model witnessed a comparable drop in performance compared to the full model.
This underlines the equal importance of these proposed training stages to the joint down-stream performance on AVSD and VisDial.
% We did not conduct ablations using either WebVid-2M or CC-3M in Stage 1 as this was sufficiently explored by other recent works \cite{vindlu} that showed the benefit of pre-training on both image and video data.

\paragraph{Effect of Pre-training Objectives \& Model Design.}
% To evaluate the effect of the newly introduced spatial-temporal objectives, 
Then, we trained a version of our model without $\mathcal{L}_\textrm{stc}$ and $\mathcal{L}_\textrm{stm}$ in Stage 1 using the same schedule and training data as our full model.
As shown in the \textit{fourth} row of \autoref{tab:ablations}, this ablated version suffered a performance drop not only in AVSD but also in VisDial.
This indicates that these losses improve not only the temporal capabilities of our model but also its spatial ones.
Then, we trained a version that sequentially applies spatial and temporal attention, as in \cite{vindlu,space_time_att}.
Since this version does not have separate spatial-temporal experts, we also omitted the previous two objectives.
As seen in the penultimate row of \autoref{tab:ablations}, this version underperformed our full model on both datasets, showcasing the effectiveness of our approach.
Finally, we trained a version without all the expert layers. 
As shown in the last row, its performance dropped compared to our full model and performed the worst on VisDial.

\begin{table}[!t]
    
    \centering
    \scalebox{0.68}[0.68]{
        
        \begin{tabular}{l cccc cc }
        \toprule
        % &
        \multirow{2}*{\textbf{Model Ablations}} & \multicolumn{4}{c}{\textbf{AVSD-DSTC7}} & \multicolumn{2}{c}{\textbf{VisDial}} \\
        
        \cmidrule(r){2-5} \cmidrule(r){6-7}
        & \textbf{B-1} & \textbf{M} & \textbf{R} & \textbf{C}
        & \textbf{R@1} & \textbf{NDCG} \\
        \midrule
         
        Full
        & $78.9$ & $31.2$ & $62.3$ & $139.8$
        & $44.2$ & $52.3$ \\
        \hdashline
        \quad w/o Tr. Stage 1
        & $76.9$ & $30.0$ & $61.4$ & $134.0$
        & $34.5$ & $44.6$ \\
        \quad w/o Tr. Stage 2
        & $77.8$ & $30.7$ & $61.7$ & $134.7$
        & $32.6$ & $44.1$ \\
        \hdashline
        \quad w/o $\mathcal{L}_\mathrm{stc}$ \& $\mathcal{L}_\mathrm{stm}$& $77.2$ & $29.9$ & $61.1$ & $133.2$
        & $33.1$ & $44.6$ \\
        \quad w/o separate $\mathcal{E}_\mathrm{spa}$ \& $\mathcal{E}_\mathrm{tmp}$
        & $77.0$ & $30.1$ & $61.1$ & $133.8$
        & $32.9$ & $43.8$ \\
        \quad w/o experts $\{\mathcal{E}_*\}$
        & $77.5$ & $30.0$ & $61.4$ & $134.8$
        & $30.6$ & $42.2$ \\
        \bottomrule                                            
\end{tabular}
    }
\caption{
    Ablation results.
        }
    \label{tab:ablations}
\end{table}
\section{Conclusion}
We presented \modelname-- a model that can jointly tackle video and visual conversational tasks using a multimodal expert-based approach that; \textit{for the first time}, disentangles the learning of the spatial and temporal features of images and videos using two separate experts.
Extensive evaluation on the respective widely used video and visual dialog datasets of AVSD and VisDial show that our model achieves new state-of-the-art zero-shot and fine-tuning performance.
Finally, we conducted the \textit{first} domain shift evaluation of AVSD and VisDial and provided insights on how to optimally leverage their respective training data.\\
-----------------------------------------------------------------------\\
% \textbf{Acknowledgment:}\\
% This work was partially funded by a LOEWE-Start-Professur (LOEWE/4b//519/05.01.002-(0006)/94), LOEWE-Spitzen-Professur (LOEWE/4a//519/05.00.002-(0010)/93), and an Alexander von Humboldt Professorship in Multimodal Reliable AI, sponsored by Germany’s Federal Ministry for Education and Research.
% \section*{Acknowledgment}
\textbf{Acknowledgment}
This work was partially funded by a LOEWE-Start-Professur (LOEWE/4b//519/05.01.002-(0006)/94), 
LOEWE-Spitzen-Professur (LOEWE/4a//519/05.00.002-(0010)/93), 
and an Alexander von Humboldt Professorship in Multimodal Reliable AI, sponsored by Germany’s Federal Ministry for Education and Research.

{
    \small
    \bibliographystyle{ieeenat_fullname}
    \bibliography{main}
}

\appendix
\section{Training Details}
\subsection{Training Objectives}
In addition to the proposed spatial-temporal contrastive learning (STC) and spatial-temporal matching (STM), we trained our model with the following established vision-language objectives. 
\paragraph{Masked Language Modeling} teaches the model to predict masked text tokens given both the visual and textual context.
As in \cite{vindlu,li2021align} we mask $15\%$ of the tokens and minimize the loss 
\begin{align}
    \mathcal{L}_\textrm{mlm} = \mathbb{E}_{(\mathbf{V}^\textrm{vis}, \mathbf{\bar{T}}^\textrm{cap})} \left[\mathcal{H}(\mathbf{y}^\textrm{mlm}, \mathbf{p}^\textrm{mlm})\right],
\end{align}
where $\mathbf{y}^\textrm{mlm}$ and $\mathbf{p}^\textrm{mlm}$ denote the ground-truth and predicted probabilities of the masked tokens whereas $\mathbf{V}^\textrm{vis}$ and $\mathbf{\bar{T}}^\textrm{cap}
$ are the visual and masked caption token embeddings, respectively.

\paragraph{Vision-Text Contrastive Learning} helps the model better align the video/image and the text features and is defined similarly to STC as
\begin{small}
\begin{equation}
    \mathcal{L}_{\textrm{vtc}} = \frac{1}{2}\mathbb{E}_{(\mathbf{V}^\textrm{vis},\mathbf{T}^\textrm{cap})}\left[\mathcal{H}\left(\mathbf{y}^\textrm{v2t}, \mathbf{p}^\textrm{v2t}\right) +  \mathcal{H}\left(\mathbf{y}^\textrm{t2v}, \mathbf{p}^\textrm{t2v}\right)\right],
\end{equation}
\end{small}

where $\mathbf{p}^\textrm{v2t}$ and $\mathbf{p}^\textrm{t2v}$ are the softmax normalized vision-to-text and text-to-vision similarities defined as in \textcolor{red}{Equation 14} and \textcolor{red}{Equation 15} of the main text. 
$\mathbf{y}^\textrm{v2t}$ and $\mathbf{y}^\textrm{t2v}$ are their respective ground-truth one-hot similarities.
\paragraph{Vision-Text Matching} is defined similarly to STM as a binary classification problem and complements the VTC by teaching the model to distinguish between matched and unmatched paired vision-text features.
We use a video/image and its corresponding caption as a positive example. The negative examples are constructed via negative sampling of captions from different visual inputs.
Formally, 
\begin{equation}
    \mathcal{L}_\textrm{vtm} = \mathbb{E}_{(\mathbf{V}^\textrm{vis}, \mathbf{T}^\textrm{cap})}\left[\mathcal{H}(\mathbf{y}^\textrm{vtm}, \mathbf{p}^\textrm{vtm})\right],
\end{equation}
where $\mathbf{p}^\textrm{stm}$ and $\mathbf{y}^\textrm{stm}$ are the predicted and the ground-truth two-class probabilities, respectively.
For completeness, we list the detailed hyperparameters of our model in \autoref{tab:hyperparams}.
\begin{table}[!t]
    \centering
    \scalebox{0.73}[0.73]{
        \begin{tabular}{llc}
        \toprule
        \textbf{Category} & \textbf{Hyperparameter}  &  \\
        \midrule
        \multirow{8}*{\textbf{Model}}
                            & Number of expert-based layers $N$ & $12$   \\
                            & Number of multimodal experts layers $L$ & $9$   \\
                            & Number of fusion experts layers $(N-L)$ & $3$ \\
                            & Joint hidden dimension $D$ & $1024$   \\
                            & Number of frames  $F$ & $4$   \\
                            & Number of patches per frame $P$ & $64$ \\
                            & Hidden dimension of LLM & $1024$   \\
                            & Dimension of LLM linear layer & $(1024, 1024)$ \\
                            & Dimension of linear layers $\Theta_*$ & $(1024, 256)$ \\

        \midrule
        \multirow{7}*{\textbf{Optimization}}
                            & Optimizer & AdamW \\
                            & Learning rate schedule & linear \\
                            & Minimum learning rate value & $5e-5$ \\
                            & Base learning rate value & $1e-4$ \\
                            & Weight decay & $0.01$ \\
                            & Gradient clipping value & $1.0$ \\
                            & Effective batch size & $48$ \\
        \midrule
        \multirow{3}*{\textbf{Hardware}}
                            & GPU model & \texttt{A100} \\
                            & Number of GPUs  & $8$ \\
                            & Distributed training &\texttt{DDP} \\
        \bottomrule
        
        \end{tabular}
    }
    \caption{Detailed hyperparameter setting of \modelname.}
    \label{tab:hyperparams}
\end{table}

\begin{table*}[!t]
    
    \centering
    \scalebox{0.83}[0.83]{
        
        \begin{tabular}{lcccccccccccccc}
        \toprule
        \multirow{2}*{\textbf{\,\,\,\,Model}}  & \multicolumn{7}{c}{\textbf{AVSD-DSTC8}} & \multicolumn{7}{c}{\textbf{AVSD-DSTC7}}\\
        \cmidrule(r){2-8} \cmidrule(r){9-15}
        & \textbf{B-1} & \textbf{B-2} & \textbf{B-3} & \textbf{B-4} & \textbf{M} & \textbf{R} & \textbf{C} &\textbf{B-1} & \textbf{B-2} & \textbf{B-3} & \textbf{B-4} & \textbf{M} & \textbf{R} & \textbf{C} \\
        %\hline
        \hline
\rowcolor{mygray} \multicolumn{15}{l}{\textit{\,\,\,\,Models from the main text}}\\
\,\,\,\,{PDC}$_\textit{\textcolor{gray}{ICLR'21}}$  \cite{le2021learning} 
& $74.9$ & $62.9$ & $52.8$ & $43.9$ & $28.5$ & $59.2$ & $120.1$
& $77.0$ & $65.3$ & $53.9$ & $44.9$ & $29.2$ & $60.6$ & $129.5$
\\
\,\,\,\,{THAM}$_\textit{\textcolor{gray}{EMNLP'22}}$ \cite{tham}  
& $76.4$ & $64.1$ & $53.8$ & $45.5$ & $30.1$ & $61.0$ & $130.4$
& $77.8$ & $65.4$ & $54.9$ & $46.8$ & {$30.8$} & {$61.9$} & $133.5$ 
\\

\,\,\,\,{DialogMCF}$_\textit{\textcolor{gray}{TASLP'23}}$ \cite{dmcf}
& $75.6$ & $63.3$ & $53.2$ & $44.9$ & $29.3$ & $60.1$ & $125.3$
& $77.7$ & $65.3$ & $54.7$ & $45.7$ & $30.6$ & $61.3$ & $135.2$

\\

$^{\vardiamondsuit}$VideoLLAMA 2$_\textit{\textcolor{gray}{arXiv'24}}$ \cite{video_llama2}
& $53.3$ & $39.0$ & $29.1$ & $22.2$ & $24.8$ & $46.3$ & $74.0$
& $56.2$ & $41.1$ & $30.7$ & $23.2$ & $26.4$ & $48.5$ & $79.2$

\\

\,\,\,\,MST-MIXER$_\textit{\textcolor{gray}{ECCV'24}}$ \cite{mst_mixer}
& $\mathbf{77.1}$     & $\mathbf{65.6}$     & $\underline{55.7}$     & $\underline{47.1}$   & $\underline{30.2}$     & $\underline{61.8}$     & $\underline{133.6}$
&$\underline{78.4}$    & $\underline{66.0}$     & $\underline{55.8}$     & $\underline{47.1}$     & $\underline{31.0}$     & $\underline{62.0}$     & $\underline{136.5}$
\\
\hline
\rowcolor{mygray} \multicolumn{15}{l}{\textit{\,\,\,\,Additional models}}\\

\,\,\,\,{MTN}$_\textit{\textcolor{gray}{ACL'19}}$ \cite{mtn}
& $-$ & $-$ & $-$ & $-$ & $-$ & $-$ & $-$ 
& $71.5$ & $58.1$ & $47.6$ & $39.2$ & $26.9$ & $55.9$ & $106.6$ 
\\
\,\,\,\,{JMAN}$_\textit{\textcolor{gray}{AAAI'20}}$ \cite{chu2020multi}     
& $64.5$ & $50.4$ & $40.2$ & $32.4$ & $23.2$ & $52.1$ & $87.5$ 
& $66.7$ & $52.1$ & $41.3$ & $33.4$ & $23.9$ & $53.3$ & $94.1$

\\

\,\,\,\,{VGD}$_\textit{\textcolor{gray}{ACL'20}}$ \cite{le-hoi-2020-video}   
& $-$ & $-$ & $-$ & $-$ & $-$ & $-$ & $-$
& $74.9$ & $62.0$ & $52.0$ & $43.6$ & $28.2$ & $58.2$ & $119.4$ 
\\

\,\,\,\,{BiST}$_\textit{\textcolor{gray}{EMNLP'20}}$ \cite{bist}              
& $68.4$ & $54.8$ & $45.7$ & $37.6$ & $27.3$ & $56.3$ & $101.7$ 
& $75.5$ & $61.9$ & $51.0$ & $42.9$ & $28.4$ & $58.1$ & $119.2$ 
\\
\,\,\,\,{SCGA}$_\textit{\textcolor{gray}{AAAI'21}}$ \cite{scga}              
& $71.1$ & $59.3$ & $49.7$ & $41.6$ & $27.6$ & $56.6$ & $112.3$
& $74.5$ & $62.2$ & $51.7$ & $43.0$ & $28.5$ & $57.8$ & $120.1$ 
\\

\,\,\,\,{RLM}$_\textit{\textcolor{gray}{TASLP'21}}$  \cite{9376902}           
& $74.6$ & $62.6$ & $52.8$ & $44.5$ & $28.6$ & $59.8$ & $124.0$ 
& $76.5$ & $64.3$ & $54.3$ & $45.9$ & $29.4$ & $60.6$ & $130.8$ 

\\

\,\,\,\,{AV-TRN}$_\textit{\textcolor{gray}{ICASSP'22}}$ \cite{av_trn}
& $-$ & $-$ & $-$ & $39.4$ & $25.0$ & $54.5$ & $99.7$
& $-$ & $-$ & $-$ & $40.6$ & $26.2$ & $55.4$ & $107.9$ 

\\

\,\,\,\,{VGNMN}$_\textit{\textcolor{gray}{NAACL'22}}$ \cite{vgnmn}      
& $-$ & $-$ & $-$ & $-$ & $-$ & $-$ & $-$ 
& $-$    & $-$    & $-$    & $42.9$ & $27.8$ & $57.8$ & $118.8$ 

\\ 

\,\,\,\,{COST}$_\textit{\textcolor{gray}{ECCV'22}}$ \cite{pham2022video}
& $69.5$ & $55.9$ & $46.5$ & $3.82$ & $27.8$ & $57.4$ & $105.1$ 
& $72.3$ & $58.9$ & $48.3$ & $40.0$ & $26.6$ & $56.1$ & $108.5$ 
\\

\,\,\,\,{MRLV}$_\textit{\textcolor{gray}{NeurIPS'22}}$ \cite{mrlv}
& $-$ & $-$ & $-$ & $-$ & $-$ & $-$ & $-$ 
& $-$ & $59.2$ & $49.3$ & $41.5$ & $26.9$ & $56.9$ & $115.9$ 
\\

\midrule
\rowcolor{ftcolor} \,\,\,\,\modelname \parbox[c]{1.5em}{\includegraphics[width=0.15in]{figures/robot_mixed.png}}
& $\underline{76.8}$     & $\underline{65.5}$     & $\mathbf{55.8}$     & $\mathbf{47.5}$   & $\mathbf{30.4}$     & $\mathbf{62.1}$     & $\mathbf{135.7}$
&$\mathbf{78.9}$    & $\mathbf{66.5}$     & $\mathbf{56.1}$     & $\mathbf{47.4}$     & $\mathbf{31.2}$     & $\mathbf{62.3}$     & $\mathbf{139.8}$
\\

\bottomrule                                            
\end{tabular}
    }
\caption{
    To complement \textcolor{red}{Table 4} of the main text, we compared our \modelname with additional fine-tuned models on AVSD-DSTC8 and AVSD-DSTC7.
   }
    \label{tab:dstc78_finetuned}
\end{table*}

\section{Additional Model Comparisons}
To complement \textcolor{red}{Table 4} of the main text, we compared our model with additional \textit{fine-tuned} baselines on the early two versions of AVSD (i.e. AVSD-DSTC8 and AVSD-DSTC7).
As shown in \autoref{tab:dstc78_finetuned}, \modelname managed to outperform these baselines as well across all metrics of the dataset.

\section{Qualitative Samples}
\begin{figure*}[!t]
    \begin{minipage}{1\linewidth}
        \centering
        \scalebox{0.97}[0.97]{
            \includegraphics[width=\textwidth]{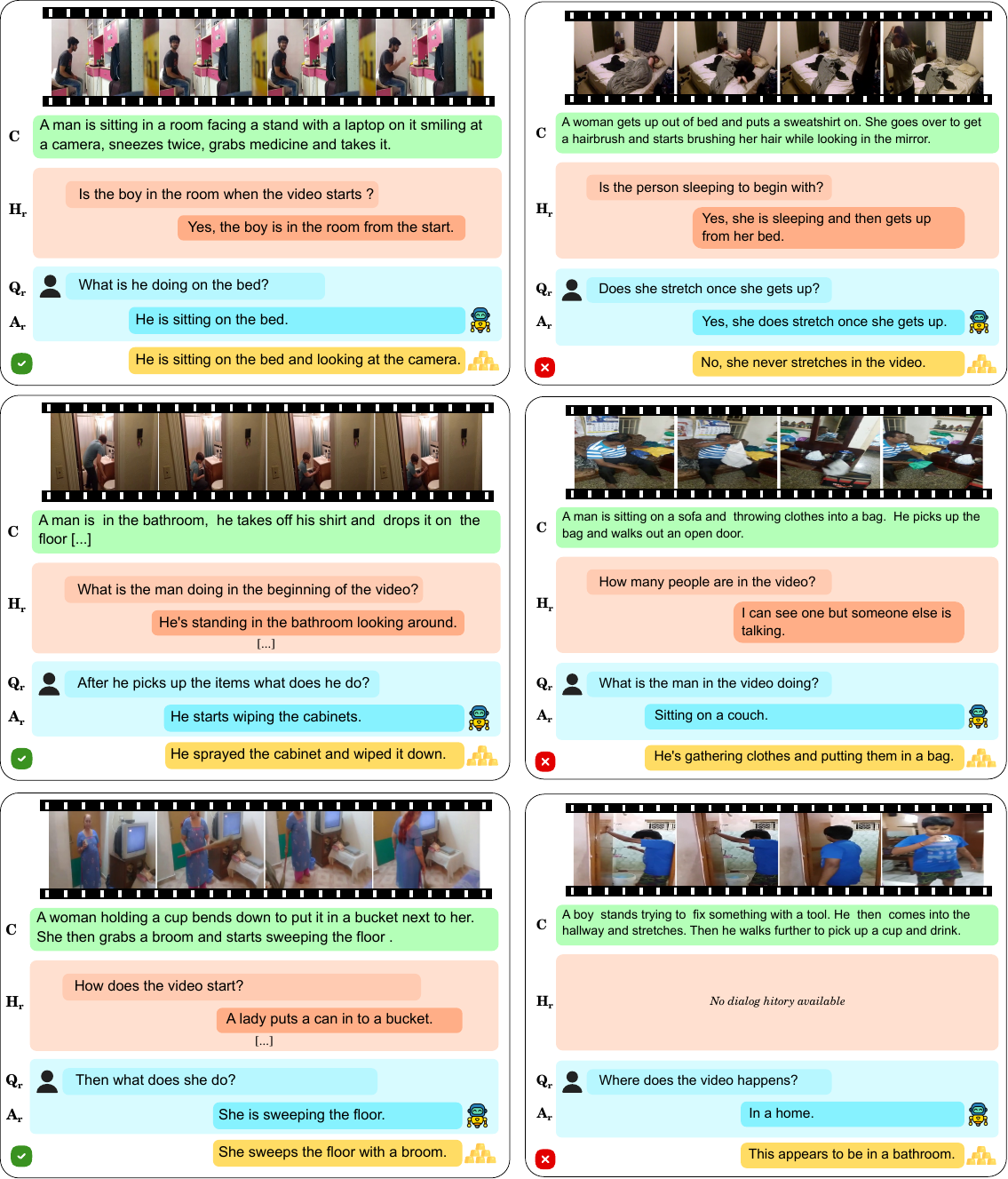}
        }
        \caption{\colorbox{qa_blue}{\textbf{Zero-shot}}
        \textbf{qualitative examples on AVSD.} We denote with C, H$_r$, Q$_r$, A$_r$ the caption, the dialog history, the current question, and its response as generated from our model, respectively. 
        (
        \img{figures/robot_mixed.png} = generated answers,
        \img{figures/gold.pdf} = golden ground-truth answers,
        \img{figures/positive.pdf} / \img{figures/negative.pdf}  = success / failure cases).
        }
        \label{fig:zs_avsd}
    \end{minipage}
\end{figure*}

\begin{figure*}[!t]
    \begin{minipage}{1\linewidth}
        \centering
        \scalebox{0.97}[0.97]{
            \includegraphics[width=\textwidth]{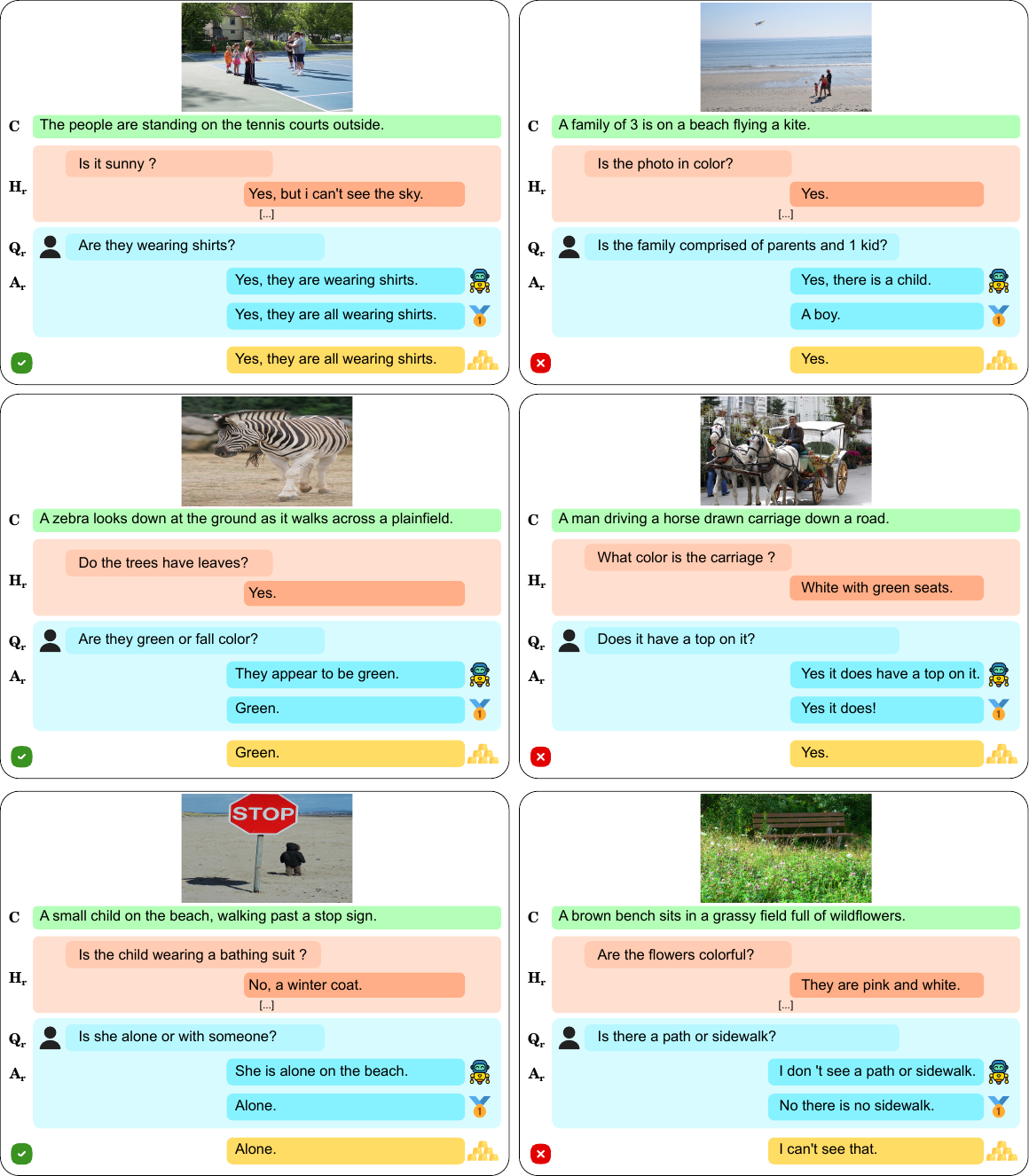}
        }
        \caption{\colorbox{qa_blue}{\textbf{Zero-shot}}
        \textbf{qualitative examples on VisDial.} We denote with C, H$_r$, Q$_r$, A$_r$ the caption, the dialog history, the current question, and its response as generated from our model, respectively. 
        (
        \img{figures/robot_mixed.png} = generated answers,
        \img{figures/top1.pdf} = top ranked candidate answers,
        \img{figures/gold.pdf} = golden ground-truth answers,
        \img{figures/positive.pdf} / \img{figures/negative.pdf}  = success / failure cases).
        }
        \label{fig:zs_visdial}
    \end{minipage}
\end{figure*}

\begin{figure*}[!t]
    \begin{minipage}{1\linewidth}
        \centering
        \scalebox{0.96}[0.96]{
            \includegraphics[width=\textwidth]{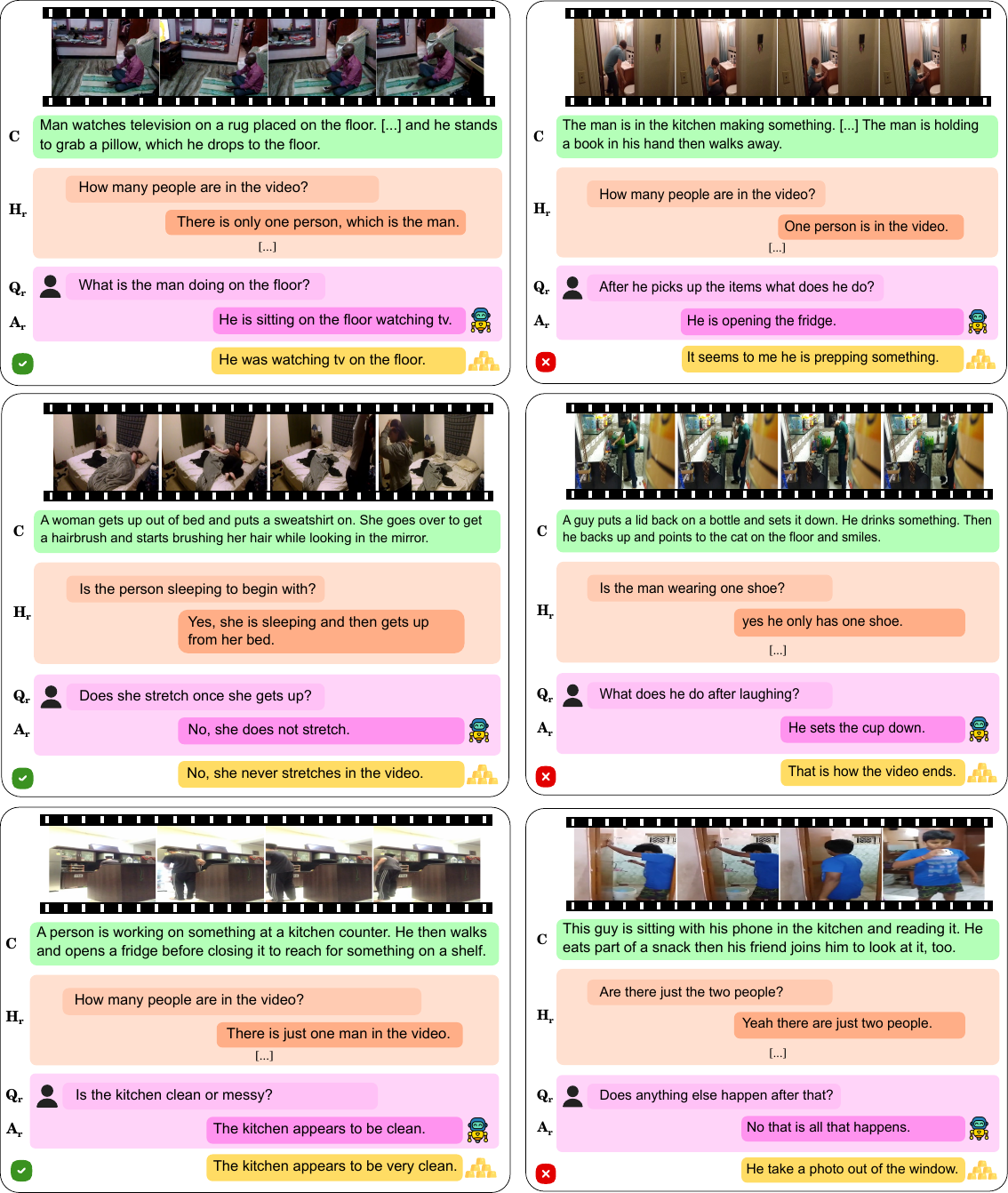}
        }
        \caption{\colorbox{qa_pink}{\textbf{Fine-tuning}} \textbf{qualitative examples on AVSD.}
        We denote with C, H$_r$, Q$_r$, A$_r$ the caption, the dialog history, the current question, and its response as generated from our model, respectively.
        (
        \img{figures/robot_mixed.png} = generated answers,
        \img{figures/gold.pdf} = golden ground-truth answers,
        \img{figures/positive.pdf} / \img{figures/negative.pdf}  = success / failure cases).
        }
        \label{fig:ft_avsd}
    \end{minipage}
\end{figure*}

\begin{figure*}[!t]
    \begin{minipage}{1\linewidth}
        \centering
        \scalebox{0.96}[0.96]{
            \includegraphics[width=\textwidth]{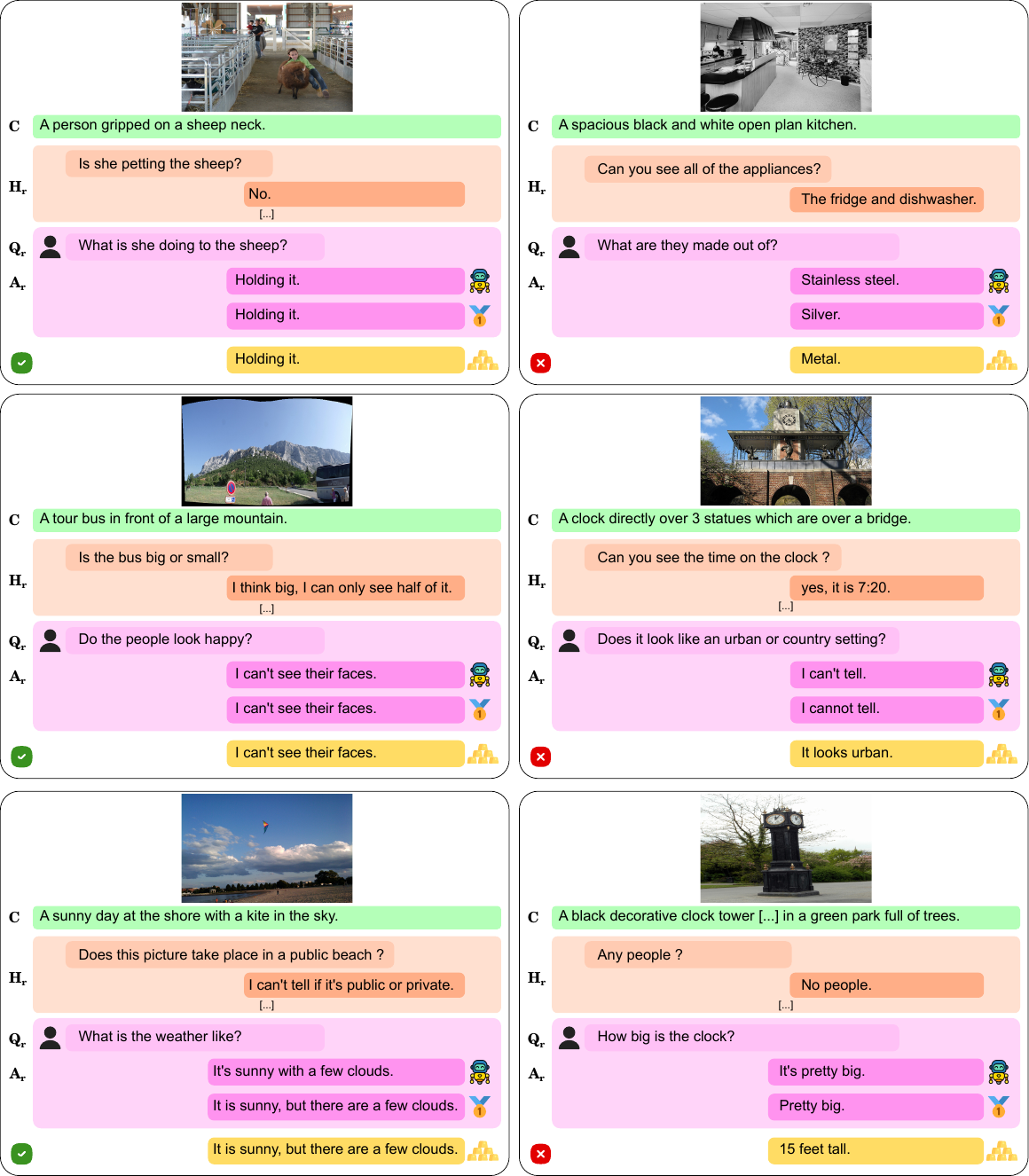}
        }
        \caption{\colorbox{qa_pink}{\textbf{Fine-tuning}} \textbf{qualitative examples on VisDial.}
       We denote with C, H$_r$, Q$_r$, A$_r$ the caption, the dialog history, the current question, and its response as generated from our model, respectively.
        (
        \img{figures/robot_mixed.png} = generated answers,
        \img{figures/top1.pdf} = top ranked candidate answers,        \img{figures/gold.pdf} = golden ground-truth answers,
        \img{figures/positive.pdf} / \img{figures/negative.pdf}  = success / failure cases).
        }
        \label{fig:ft_visdial}
    \end{minipage}
\end{figure*}

We provide additional qualitative samples comprising of both success and failure cases of our model.
\autoref{fig:zs_avsd} and \autoref{fig:zs_visdial} illustrate some zero-shot samples for AVSD and VisDial, respectively. 
Additional fine-tuning examples for both datasets are shown in \autoref{fig:ft_avsd} and \autoref{fig:ft_visdial}.

As defined in \textcolor{red}{Section 3.1} of the main text, 
we denote with C, H$_r$, and Q$_r$ the caption, the dialog history, and the current question, respectively. 
Similar to \textcolor{red}{Figure 5} of the main text, we highlight the caption in \colorbox{c_green}{green}, the dialog history in \colorbox{h_orange}{orange}, and the current question-answer pair in \colorbox{qa_blue}{blue} for zero-shot and \colorbox{qa_pink}{pink} for fine-tuning evaluation.
Furthermore, we use the symbols \img{figures/robot_mixed.png} and \img{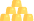} to indicate the generated and the golden ground-truth answers, respectively.
\img{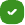} / \img{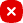} mark success / failure cases.
For VisDial, we additionally use \img{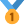} to show the top ranked candidate answers (i.e. the most similar to the generated responses).

\end{document}